\documentclass[acmsmall,screen]{acmart}
\usepackage{booktabs}
\usepackage{tabularx}
\usepackage{multirow}
\usepackage{colortbl}
\usepackage{xcolor}
\usepackage{tikz}
\usetikzlibrary{shapes.geometric,positioning,arrows.meta,calc,decorations.pathreplacing}
\usepackage{fontawesome5}
\usepackage{soul}
\usepackage[most]{tcolorbox}
\usepackage{pifont}
\newcommand{\cmark}{{\color{green!55!black}\ding{51}}}
\newcommand{\xmark}{{\color{red!75!black}\ding{55}}}
\newcommand{\hib}{{\color{green!55!black}$\boldsymbol{\uparrow}$}}

\newcommand{\pmark}{{\color{orange!85!black}$\boldsymbol{\sim}$}}
\newtcolorbox{keypoint}{enhanced, breakable, colback=green!6, colframe=green!55!black, boxrule=0pt, leftrule=3.5pt,
  arc=0pt, outer arc=0pt, left=7pt, right=7pt, top=5pt, bottom=5pt, before skip=10pt, after skip=10pt}
\usepackage{fvextra}

\definecolor{navy}{HTML}{1F3A68}
\definecolor{gold}{HTML}{D9A62E}
\colorlet{hlbad}{red!22}
\colorlet{hlmiss}{yellow!55}
\colorlet{hlfix}{green!25}
\tcbset{exbox/.style={enhanced, nobeforeafter, box align=top, arc=2pt, boxrule=0.8pt, left=3pt, right=3pt, top=2pt,
  bottom=2pt, fonttitle=\sffamily\bfseries\footnotesize, fontupper=\footnotesize, fontlower=\scriptsize,
  coltitle=black, colbacktitle=white, toptitle=1pt, bottomtitle=1pt, segmentation style={dashed}}}

\setcopyright{none}
\acmJournal{TODAES}

\begin{document}

\title[VeriLadder]{VeriLadder: A Multi-Agent Specification Refinement System for Tool-Verified Verilog Datasets}

\author{Saeid~Rajabi}
\affiliation{%
  \institution{University of Delaware}
  \city{Newark}
  \state{Delaware}
  \country{USA}}
\author{Chengmo~Yang}
\affiliation{%
  \institution{University of Delaware}
  \city{Newark}
  \state{Delaware}
  \country{USA}}
\author{Satwik~Patnaik}
\affiliation{%
  \institution{University of Delaware}
  \city{Newark}
  \state{Delaware}
  \country{USA}}

\renewcommand{\shortauthors}{Rajabi et al.}

\begin{abstract}
Training data is a bottleneck in AI, especially for large language models (LLMs) that write Verilog, because register-transfer-level (RTL) code is rarely published and existing datasets mostly check code for syntax, not instruction correctness.
This paper presents VeriLadder, a five-level Verilog dataset, and the multi-agent framework that builds it, in which LLM agents write and revise specifications, and EDA tools (a synthesizer and a formal equivalence checker) decide what is accepted. Its levels follow an engineer's checks.
Level~1 labels designs from two large Verilog corpora, namely RTLCoder and PyraNet, with synthesis results, failure causes, critical warnings and post-synthesis metrics. 28\% of RTLCoder designs fail synthesis and are kept as negatives.
Level~2 grades each instruction on four specification qualities, and low grades predict failed rebuilds.
Level~3 rewrites each instruction into a refined specification and accepts it only if an LLM from another model family, reading only that specification, writes a circuit formally equivalent to the original. This raises the share of rebuildable designs from 26.1\% to 78.2\%, keeping every failed specification with its counterexample.
Levels~4 and~5 add verified paraphrases and one-fact variants, hard negatives that differ only in one behavior from the verified design. VeriLadder labels the synthesizability of 716,627 designs and, from RTLCoder, provides 11,247 refined specifications, 32,393 paraphrases and 8,397 one-fact variants.
Fine-tuning Qwen2.5-Coder-7B-Instruct on VeriLadder raises functional pass@1 from 31.7\% to 50.0\% on VerilogEval~v2 and from 35.0\% to 45.2\% on RTLLM~v2. The dataset and fine-tuned models are publicly available.

\smallskip\noindent\raisebox{-0.15em}{\includegraphics[height=0.95em]{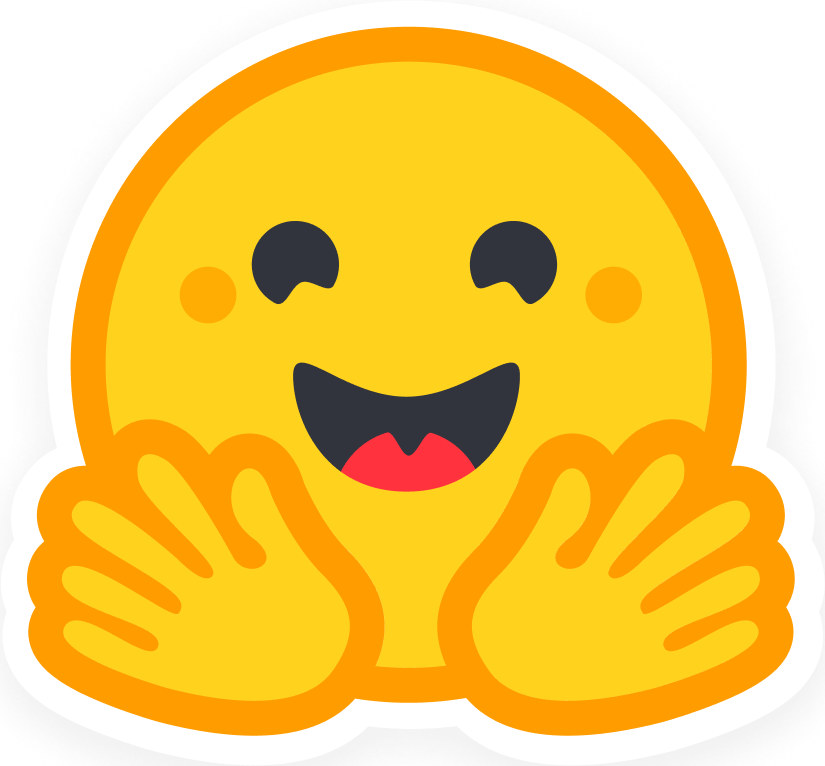}}~\url{https://huggingface.co/collections/saeidrajabi/veriladder}\footnote{The prompts and the pipeline will be released upon acceptance.}
\end{abstract}

\begin{CCSXML}
<ccs2012>
<concept><concept_id>10010583.10010682.10010689</concept_id><concept_desc>Hardware~Hardware description languages and compilation</concept_desc><concept_significance>500</concept_significance></concept>
</ccs2012>
\end{CCSXML}
\ccsdesc[500]{Hardware~Hardware description languages and compilation}

\keywords{Verilog, RTL generation, large language models, dataset, logic synthesis, specification, formal equivalence checking, paraphrase robustness, contrastive learning}

\maketitle

\section{Introduction}
\label{sec:intro}

Modern chip design relies on computer-aided design (CAD) tools, also called electronic design automation (EDA) tools.
Designers describe a digital circuit at the register-transfer level (RTL), usually in the Verilog language, and CAD tools turn this description into gates, check it, and finally lay it out on silicon.
Writing and checking the RTL is still mostly manual work, and it takes a large share of the effort of a chip design project, so tools that help with this step are in high demand.
Large language models (LLMs) are a promising such tool. They can write Verilog from a description in natural language~\cite{blocklove2023chipchat,chang2023chipgpt}, they have been adapted to chip-design tasks in industry~\cite{liu2023chipnemo}, and they are now studied for many steps of the design flow~\cite{pan2025llmeda,zhong2023llm4eda}.
Several open-source models have been trained specifically to generate RTL~\cite{liu2025rtlcoder,zhu2025codevr1,nadimi2025pyranet}, and benchmarks such as VerilogEval and RTLLM measure how well they do~\cite{liu2023verilogeval,lu2024rtllm}.

For such models, training data is one of the main bottlenecks.
The performance of LLMs improves predictably with both model size and the amount of training data~\cite{kaplan2020scaling,hoffmann2022chinchilla}. At the same time, the supply of public human-written text is expected to run out~\cite{villalobos2022rundata}, and carefully selected data can matter more than a larger amount of data~\cite{gunasekar2023textbooks}, which has turned attention from models to data~\cite{zha2023datacentric}.
Hardware is an extreme case.
RTL code is valuable intellectual property and is rarely published, so public Verilog is scarce compared with software code, and its quality is uneven.
Progress of LLMs in hardware design therefore depends on making the limited public data as useful as possible, by checking it, labeling it and repairing what is wrong.

The input to an RTL model is an instruction in natural language, and the model learns from datasets of instruction--code pairs, in which the instruction asks for a design and the code is a Verilog module that implements it. The instruction is the only description of the task, so it serves as the \emph{specification} of the design.
Several such datasets have been built (Section~\ref{sec:background}), but they check their samples only in part.

This leaves a gap between what the datasets contain and what a model needs to learn.
A hardware engineer checks a design in a fixed order.
First, the code must be able to become hardware at the end, which means that a \emph{synthesis tool} must translate it into a circuit of logic gates and flip-flops.
Second, the specification must describe what the circuit does precisely enough that another engineer would build functionally the same circuit from it. Such a specification \emph{determines} the circuit, and only then can anyone decide whether a generated design is correct with respect to the specification.
Existing datasets check neither property well, and they leave six gaps.
\begin{itemize}
  \item[(i)] \emph{Synthesis is not checked.} Datasets check the code only for syntax or compilation, which does not show that it becomes hardware. Designs that fail are either filtered out or kept without a label, so the root cause of each failure is lost, and designs that synthesize only with critical warnings are not marked.
  \item[(ii)] \emph{Specification quality is not measured.} No dataset measures how well an instruction specifies its design.
  \item[(iii)] \emph{Failing specifications are not repaired.} A few datasets test, in a \emph{round trip}, whether another LLM that reads only the specification writes code equivalent to the paired code, but when it cannot, they discard the pair or keep it with a label, and none repairs the specification until it describes the circuit precisely.
  \item[(iv)] \emph{One wording per design.} Each design comes with a single wording, although LLMs are sensitive to how a prompt is worded~\cite{wang2023recode} and, in real-world settings, different engineers ask for the same design in different words.
  \item[(v)] \emph{No functional hard negatives.} A model learns which details of a specification matter from \emph{hard negatives}, examples that are almost correct but differ from the correct design in one behavior. In existing datasets, the only negative examples are code that fails to compile or contains a deliberately inserted bug, not code that compiles and implements a nearby specification, such as the same counter with a different reset value.
  \item[(vi)] \emph{No data for learning to refine.} Open-source models are increasingly trained on the outputs of a stronger teacher LLM, and CodeV-R1 and VeriThoughts release a teacher's reasoning for writing code from a description~\cite{zhu2025codevr1,yubeaton2025verithoughts}; no dataset teaches the step that comes before, turning an imprecise request into a precise specification, which is the work a hardware engineer does before writing RTL.
\end{itemize}
As a result, a model trained on these datasets learns from code that does not become hardware and from specification-to-RTL translations that are not correct, because the instruction contradicts the code or is satisfied by many functionally different circuits.

\begin{figure*}[t]
\centering
\includegraphics[width=\linewidth]{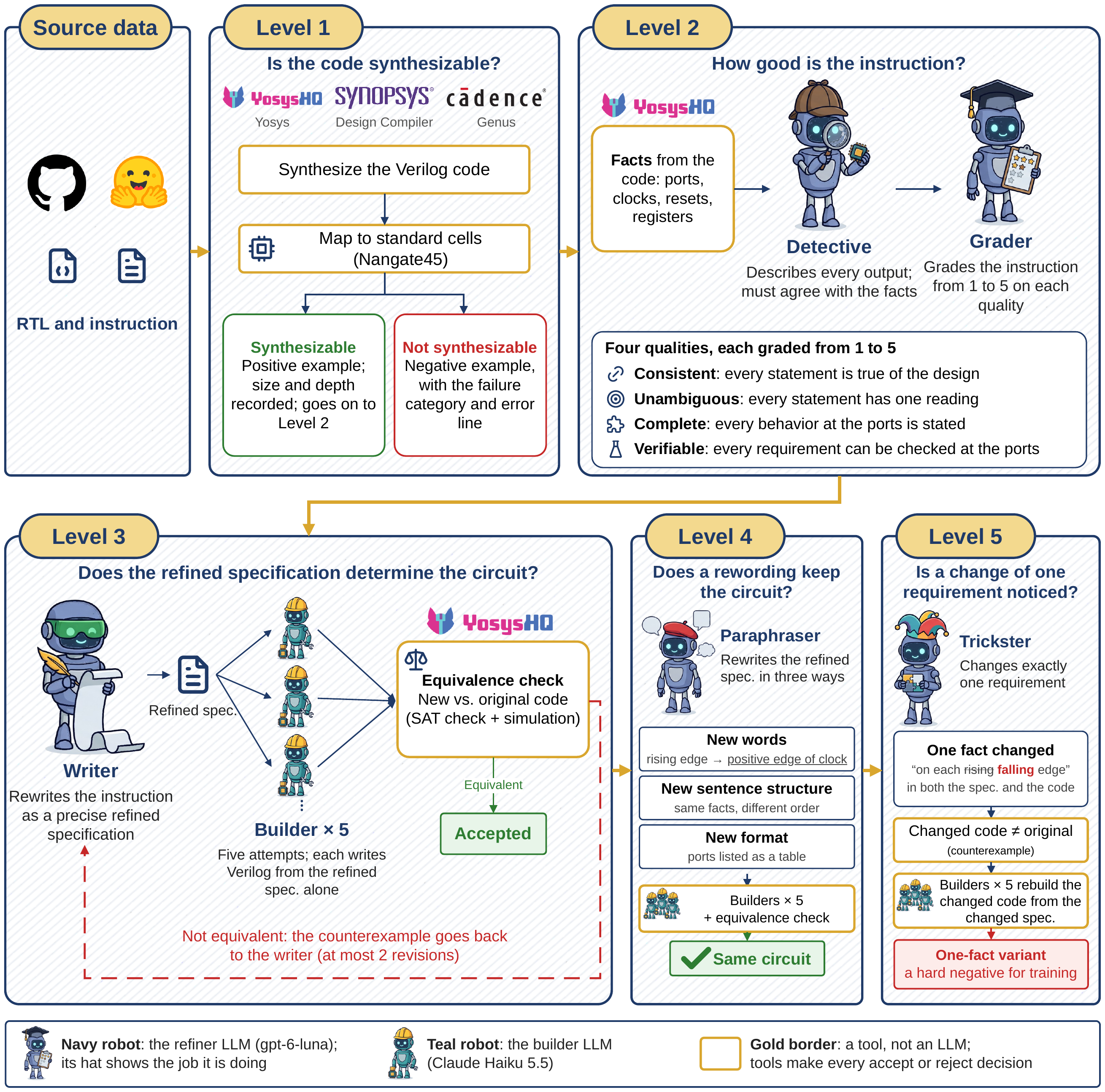}
\caption{Overview of VeriLadder. The navy robots are roles of one LLM, the \emph{refiner} (the hat shows the role), the teal robot is a second LLM, the \emph{builder}, and boxes with a gold border are tools. \textbf{Level 1} labels each design as synthesizable or not and records why it fails. \textbf{Level 2} grades the instruction on four qualities, using facts extracted by Yosys. \textbf{Level 3} rewrites it as a refined specification, accepted only if code written from it alone is verified equivalent to the original, and returns each counterexample to the writer. \textbf{Levels 4 and 5} add paraphrases and one-fact variants, verified with the same check.}
\label{fig:framework}
\end{figure*}

To bridge this gap, this paper proposes VeriLadder, a framework for \emph{multi-agent, tool-verified specification refinement}, and the Verilog dataset it builds, whose labels follow the engineer's order in five \emph{levels} (Figure~\ref{fig:framework}).
Several LLM agents with separate roles describe, grade, rewrite, rebuild and vary each design, and tools check every result and return their feedback, such as a counterexample, to the agent that wrote it.
The roles are kept separate on purpose. An LLM that checks its own specification can fill its gaps from what it knows about the code, and it tends to favor its own writing~\cite{panickssery2024selfpref}. An independent LLM, the \emph{builder}, sees only the specification, so it can rely only on what the specification states.
The LLMs only write and revise specifications, and whether a design or a specification is accepted is always decided by a tool, either a synthesis tool or a formal equivalence checker.
VeriLadder addresses all six aforementioned gaps. Its five levels label whether the code becomes hardware~(i), grade how well the instruction specifies it~(ii), refine the instruction until it determines the circuit~(iii), and add verified paraphrases~(iv) and one-fact variants that differ from the design in exactly one behavior~(v). Refinement trajectories stored at Level~3 address gap~(vi).

% \paragraph{Contributions.}
Our \textbf{contributions} are summarized as follows.
\begin{itemize}
  \item \textbf{A multi-agent framework for dataset refinement.} VeriLadder performs multi-agent, tool-verified specification refinement and is, to our knowledge, the first framework that repairs failing specifications and verifies every repair by formal equivalence. LLM agents grade and rewrite each instruction, and a rewrite is accepted only if an independent LLM, reading only that specification, builds a circuit that is formally equivalent to the original. Refinement raises the share of designs rebuilt as an equivalent circuit from 26.1\% to 78.2\%, also with an open-source builder (28\% to 71\%), and stores every \emph{refinement trajectory}, the sequence of rewrites of an instruction with the counterexample that rejected each, as data for learning to refine (Section~\ref{sec:level3}).
  \item \textbf{Metrics for the quality of hardware specifications as data.} Four rubric-based metrics (consistency, unambiguity, completeness and verifiability) predict whether a specification determines its circuit (area under the ROC curve 0.89), and LLMs from different model families give nearly the same grades (Section~\ref{sec:level2}).
  \item \textbf{Verified paraphrases.} 32,393 paraphrases of the refined specifications, each verified by the same round trip to describe the same circuit, for training or testing robustness of the RTL code generation model to the prompt variations (Section~\ref{sec:level4}).
  \item \textbf{One-fact variants as hard negatives.} 8,397 variants whose specification and code differ from the design in exactly one behavior, each with a counterexample and a check that the changed specification determines the changed code, for preference optimization and contrastive learning (Section~\ref{sec:level5}).
\end{itemize}

\section{Background and Related Work}
\label{sec:background}
\subsection{Training data for code LLMs}
LLMs perform well at generating software code in languages such as Python, C, and C++, largely because they are trained on large amounts of public repositories.
Fine-tuning GPT-3 on public GitHub code produced Codex, which solved 28.8\% of the HumanEval problems while GPT-3 solved almost none, and a version of which powers GitHub Copilot~\cite{chen2021codex}.
Later work showed that the quality of such data can matter more than its size. For example, phi-1, a 1.3B-parameter model trained on only 7 billion tokens of carefully selected, textbook-like code, solves 50.6\% of HumanEval, more than code generation models ten times larger (in terms of parameter size) that were trained on over a hundred times more tokens~\cite{gunasekar2023textbooks}.
Hardware lacks this advantage, because RTL code is mostly proprietary and public Verilog is scarce.

\begin{table}[t]
\centering\small
\caption{Comparison of VeriLadder with prior Verilog datasets, by the checks they apply and the training data they provide. PPA is power, performance and area after synthesis.}
\label{tab:related}
\resizebox{\linewidth}{!}{%
\begin{tabular}{lr cccc cccc ccc}
\toprule
 & & \multicolumn{4}{c}{Code} & \multicolumn{4}{c}{Specification} & \multicolumn{3}{c}{Added training data} \\
\cmidrule(lr){3-6}\cmidrule(lr){7-10}\cmidrule(lr){11-13}
Dataset & Samples & Syntax & Synthesis & Failure & PPA & Quality & Checked & Formal & Repaired & Teacher & Para- & Hard \\
 & & & & cause & & metric & vs.\ code & check & & trajectories & phrases & negatives \\
\midrule
RTLCoder~\cite{liu2025rtlcoder} & $\sim$27K & \cmark & \xmark & \xmark & \xmark & \xmark & \xmark & \xmark & \xmark & \xmark & \xmark & \xmark \\
MG-Verilog~\cite{zhang2024mgverilog} & 11K & \cmark & \xmark & \xmark & \xmark & \xmark & \xmark & \xmark & \xmark & \xmark & \xmark$^{e}$ & \xmark \\
PyraNet~\cite{nadimi2025pyranet} & 692K & \cmark & \xmark & \pmark$^{a}$ & \xmark & \pmark$^{b}$ & \xmark & \xmark & \xmark & \xmark & \xmark & \xmark \\
MetRex~\cite{abdelatty2025metrex} & 25.9K & \cmark & \cmark & \xmark & \cmark & \xmark & \xmark & \xmark & \xmark & \xmark & \xmark & \xmark \\
DeepCircuitX~\cite{li2025deepcircuitx} & 140K files & \cmark & \cmark & \xmark & \cmark & \pmark$^{c}$ & \xmark & \xmark & \xmark & \xmark & \xmark & \xmark \\
CodeV-R1~\cite{zhu2025codevr1} & 87K & \cmark & \cmark & \xmark & \xmark & \xmark & \cmark & \xmark & \xmark & \cmark$^{d}$ & \xmark & \xmark \\
VeriThoughts~\cite{yubeaton2025verithoughts} & 20.2K & \cmark & \cmark & \xmark & \xmark & \xmark & \cmark & \cmark & \xmark & \cmark$^{d}$ & \xmark & \xmark \\
\midrule
VeriLadder & 717K / 11K$^{g}$ & \cmark & \cmark & \cmark & \cmark & \cmark & \cmark & \cmark & \cmark & \cmark$^{f}$ & \cmark & \cmark \\
\bottomrule
\end{tabular}}
\par\smallskip\footnotesize \cmark{} yes, \xmark{} no, \pmark{} partly.
$^{a}$Only designs with missing dependencies are labeled.
$^{b}$An LLM rates the code, not the description.
$^{c}$Humans grade a sample of the descriptions.
$^{d}$Reasoning traces of DeepSeek-R1 for writing code from a description.
$^{e}$Descriptions at different levels of granularity, not paraphrases.
$^{f}$Every refined specification with its counterexample, from the original instruction to the accepted specification.
$^{g}$Designs with Level~1 labels / designs with a verified refined specification.
\end{table}

\subsection{Verilog datasets}
The success of code LLMs in software has motivated their use for hardware design. LLMs can write Verilog from a description~\cite{blocklove2023chipchat,chang2023chipgpt}, and models such as RTLCoder, PyraNet and CodeV-R1 have been trained specifically to generate RTL~\cite{liu2025rtlcoder,nadimi2025pyranet,zhu2025codevr1}.
Such models need Verilog training data, and public RTL code is scarce, so researchers have built dedicated datasets in several ways.
VeriGen collected Verilog files from GitHub and textbooks~\cite{thakur2023benchmarking,thakur2024verigen}.
RTLCoder asked GPT-3.5 to write both the instruction and the code, giving about 27K pairs~\cite{liu2025rtlcoder}.
MG-Verilog, PyraNet, CodeV-R1 and VeriThoughts collected code and asked an LLM to describe it~\cite{zhang2024mgverilog,nadimi2025pyranet,zhu2025codevr1,yubeaton2025verithoughts}, and MetRex and DeepCircuitX added area, delay or power after synthesis~\cite{abdelatty2025metrex,li2025deepcircuitx}.
Most of these datasets consist of instruction--code pairs.
The instruction plays the role of a \emph{specification}, a text that states what a module must do, such as its ports, its behavior, and its response to clock and reset. Usually it does so in an informal and incomplete way, and this paper calls it the \emph{original instruction}.

\subsection{Quality of Verilog datasets}
The datasets differ most in how they check their samples, first the code and then the specification (Table~\ref{tab:related}).
For the code, RTLCoder keeps only pairs that pass a syntax check~\cite{liu2025rtlcoder}, PyraNet asks an LLM to rank the code of each sample from 0 to 20 for coding style and efficiency~\cite{nadimi2025pyranet}, and MetRex synthesizes every design and asks an LLM to repair those that fail~\cite{abdelatty2025metrex}.
A syntax check or an LLM rating does not show that the code becomes hardware, and a repair removes the evidence of why a design failed. No dataset grades all of its specifications either, and the LLM rating of PyraNet judges the code, not its specification.

For the specification, the question is whether it \emph{determines} its circuit, that is, whether every circuit that satisfies it is functionally identical to the paired code.
Only then can a model trained on the pair learn the mapping from text to circuit, and only then can a generated design be judged against the text.
VerilogEval keeps an LLM-written problem description if at least one sampled solution passes simulation~\cite{liu2023verilogeval}.
The closest works to VeriLadder are CodeV-R1~\cite{zhu2025codevr1} and VeriThoughts~\cite{yubeaton2025verithoughts}, which both use a round trip and release the reasoning of a teacher model.
CodeV-R1 collects about 150K Verilog snippets from GitHub, asks DeepSeek-V3 to describe each one, and asks DeepSeek-R1 to write code from the description.
A rule-based testbench simulates the rebuilt and the original code on random inputs and keeps a pair only if no output differs, but only for its reinforcement-learning data; the 87K pairs used for fine-tuning are not checked against the original code.
VeriThoughts asks one LLM to describe each design, asks DeepSeek-R1 to write code from the description, and labels the pair as consistent if a formal check bounded to 50 cycles finds no difference; inconsistent pairs are kept with their label.
Both only judge a specification as it is. A specification that fails is kept, labeled or discarded, but never repaired, so the designs behind it are lost or keep an imprecise text, and no dataset shows how a failing specification can be made precise.
VeriRL rewrites every PyraNet description it uses into a structured problem statement with ports, timing and behavior, but the rewrite is not checked against the code, so it may still allow other circuits~\cite{teng2025verirl}.
Table~\ref{tab:related} compares the checks and the data of each prior dataset with those of VeriLadder.

\section{Building VeriLadder}
\label{sec:construction}
\begin{figure}[t]
\centering
\includegraphics[width=\linewidth]{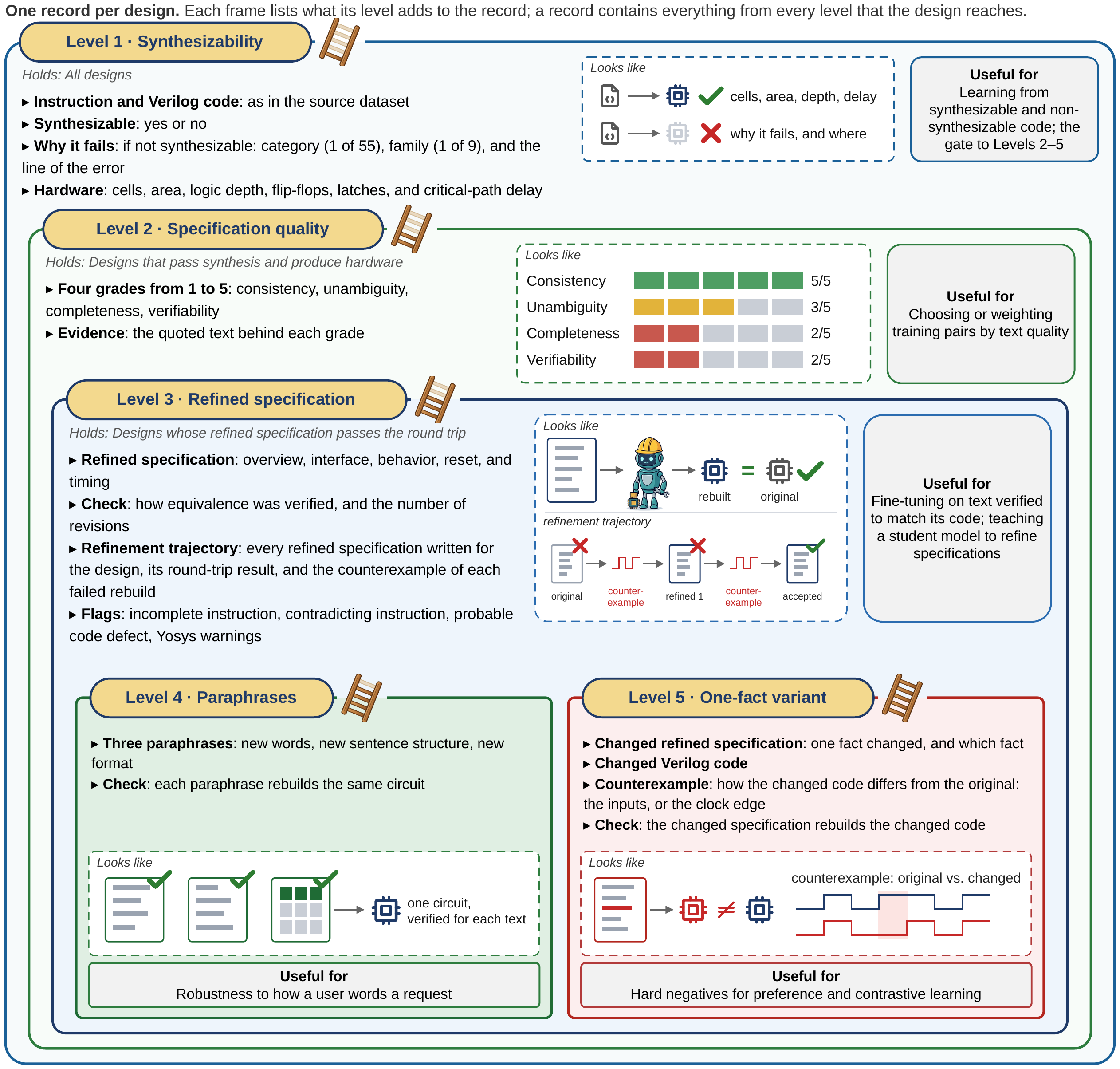}
\caption{What a record of VeriLadder contains. Each frame lists what one level adds and holds only the designs that reach that level. At Level~3 the record keeps the refinement trajectory, and at Level~5 the counterexample of every one-fact variant. ``Looks like'' boxes are simplified drawings, ``Useful for'' names a use of what the level adds, and Figures~\ref{fig:ex-level1}--\ref{fig:ex-level5} show real records.}
\label{fig:overview}
\end{figure}

This section describes how VeriLadder is built. It first introduces the source designs (Section~\ref{sec:source}) and then the five levels in order, synthesizability (Section~\ref{sec:level1}), specification quality (Section~\ref{sec:level2}), refined specifications (Section~\ref{sec:level3}), paraphrases (Section~\ref{sec:level4}) and one-fact variants (Section~\ref{sec:level5}).
The levels form a ladder, because each level works only on the designs that the levels before it kept and asks a stricter question about them. The dataset stores one record per design, which contains the original instruction and code together with the labels and texts that each level adds, up to the highest level the design reaches (Figure~\ref{fig:overview}).
Levels~2--5 are carried out by the VeriLadder framework (Figure~\ref{fig:framework}), whose agents are roles of two LLMs. The \emph{refiner} (gpt-6-luna) describes, grades, rewrites, paraphrases and varies the specifications, and the \emph{builder} (Claude Haiku~5.5) writes code from them.

\subsection{Source designs}
\label{sec:source}
VeriLadder is built from the RTLCoder dataset~\cite{liu2025rtlcoder}, instruction--code pairs written by GPT-3.5.
To test whether the Level-1 findings hold beyond one dataset, Level~1 is also applied to PyraNet~\cite{nadimi2025pyranet}.
To prevent benchmark contamination, every design whose instruction has a Rouge-L F1 above 0.5 with the description of a VerilogEval or RTLLM problem~\cite{liu2023verilogeval,lu2024rtllm}, or whose code has a Rouge-L F1 above 0.8 with a reference solution, is removed before any level. This leaves 26,076 RTLCoder and 690,551 PyraNet designs, and all statistics in Sections~\ref{sec:construction} and~\ref{sec:stats} refer to them.
% Both datasets are public, and VeriLadder adds labels and text to them but no new designs.
The approach is general and can be applied beyond a single dataset.
Levels~2--5 are applied to the 15,656 effectively synthesizable RTLCoder designs (Section~\ref{sec:level1}), of which 14,385 pass two entry checks, since Yosys must extract the facts that Level~2 needs and the equivalence checker must verify the design as equivalent to itself.
Every request to an LLM and every response is stored, together with the prompts and the code of the run, and only public data is sent to commercial LLM services.

%% ------------------------------------------------------------
\subsection{Level 1: Synthesizability}
\label{sec:level1}
Level~1 asks whether the code of a design becomes hardware at the end of the chip design flow.
Figure~\ref{fig:ex-level1} shows one design for each outcome.
% Real records for Levels 1-3. Example design: RTLCoder id 53 (digital_system, batch 9), combinational; it matches no
% VerilogEval or RTLLM problem (Rouge-L below the exclusion thresholds). Level 1: ids 53 (passes, no critical warning),
% 16711 (altiobuf_in, fails), 1428 (power_good, passes with a critical warning).
% Level-1 numbers: paper/stats/cache/labels.parquet and cross_corpus_taxonomy/runs/rtlcoder_timing/timing.jsonl.
\begin{figure}[tp]
\centering
\scriptsize
\noindent
\begin{tcolorbox}[exbox, fontupper=\scriptsize, colback=green!4, colframe=green!45!black, width=0.32\linewidth, title={\texttt{digital\_system}: synthesizable}, equal height group=lv1]
\raggedright
\begin{Verbatim}[fontsize=\fontsize{6.4}{7.6}\selectfont]
always @* begin
  if ((a+b+c+d)%3 == 0)
    out <= 1;
  else out <= 0;
end
\end{Verbatim}
Synthesizable: \cmark{} yes\\
Critical warnings: none\\
Cells: 5 \quad Area: 5.9\,\textmu m$^2$\\
Flip-flops: 0 \quad Logic depth: 3\\
Critical-path delay: 0.11\,ns
\tcblower
A small combinational circuit: a positive training example.
\end{tcolorbox}\hfill
\begin{tcolorbox}[exbox, fontupper=\scriptsize, colback=red!3, colframe=red!60!black, width=0.32\linewidth, title={Design 16711: not synthesizable}, equal height group=lv1]
\raggedright
\begin{Verbatim}[fontsize=\fontsize{6.4}{7.6}\selectfont]
always @ (posedge datain
          or negedge datain)
  buffer <= datain;
\end{Verbatim}
Synthesizable: \xmark{} no\\
Yosys: ``ERROR: Multiple edge sensitive events found for this signal!''\\
Family: clocking and sequential inference
\tcblower
Valid Verilog that simulates, but no flip-flop reacts to both edges of a signal.
\end{tcolorbox}\hfill
\begin{tcolorbox}[exbox, fontupper=\scriptsize, colback=orange!4, colframe=orange!70!black, width=0.32\linewidth, title={Design 1428: critical warning}, equal height group=lv1]
\raggedright
\begin{Verbatim}[fontsize=\fontsize{6.4}{7.6}\selectfont]
or  or0  (pgood_out, A2, A1);
or  or1  (pgood_out, B2, B1);
not not0 (pgood_out, pgood_out);
\end{Verbatim}
Synthesizable: \cmark{} yes\\
Critical warning: conflicting drivers (``Warning: multiple conflicting drivers for power\_good...'')
\tcblower
Three gates drive the wire \texttt{pgood\_out}; a pass label alone would hide the defect.
\end{tcolorbox}
\caption{Real records of Level~1, one for each outcome of the synthesis label. A synthesizable design with its post-synthesis metrics, a design that is not synthesizable with the Yosys error and its family, and a synthesizable design with a critical warning.}
\label{fig:ex-level1}
\end{figure}

A design is \emph{synthesizable} if a synthesis tool, here Yosys~\cite{wolf_yosys}, translates it into a gate-level netlist without errors, and a warning is \emph{critical} if it points to a likely hardware defect. The netlist is then mapped to the cells of a standard-cell library (here, the Nangate 45\,nm library), which gives the \emph{cell count}, the \emph{area} and the \emph{logic depth}, the largest number of cells on any path from an input port or a flip-flop to an output port or a flip-flop. Static timing analysis of the mapped netlist gives the \emph{critical-path delay}, the longest delay on any such path, which limits the clock frequency.

\subsubsection{Why synthesis comes first}
RTL code that cannot be synthesized never becomes hardware. The later levels also need the synthesized circuit, not only the source code, because the scorer of Level~2 reads facts that Yosys extracts from the synthesized design and the equivalence checker of Level~3 compares two synthesizable circuits. Level~1 also keeps the designs that fail, labeled with the cause and the line of the failure, so that they can serve as negative examples for preference optimization, which trains a model toward preferred and away from dispreferred outputs~\cite{ouyang2022rlhf}. Only designs that synthesize into at least one standard cell, called \emph{effectively synthesizable}, continue to Level~2, because a design with no cells has no logic for the later levels to check.

\subsubsection{Synthesis outcome}
Each design is elaborated and synthesized into a generic gate-level netlist with Yosys under a fixed time budget (the Level~1 panel of Figure~\ref{fig:framework}).
The outcome is \emph{synthesizable}, \emph{not synthesizable}, or \emph{timeout}.

\subsubsection{Root cause of a failure}
A design that fails is annotated with the root cause of its failure.
The error that the synthesizer reports is classified by its type into 55 categories, which are grouped into nine families by the kind of defect. The families are unresolved references (a module, signal or file that does not exist), syntax, width and constant evaluation, clocking (for example, a process that reacts to both edges of a signal), unsupported constructs, simulation-only code, combinational loops that stop synthesis, timeouts, and unclassified.
The failure is also localized to the line of code that the synthesizer reports.
The family, the category and the line together state what kind of defect the design has, what exactly it is, and where it is.

\subsubsection{Critical warnings}
A design can synthesize and still contain a defect that the synthesizer only warns about.
Level~1 flags seven such defects as \emph{critical warnings}. They are conflicting drivers on a net, a wire that is used but never driven, an inferred latch, a combinational loop that Yosys reports only as a warning, an implicitly declared signal, an out-of-range bit select, and an \texttt{initial} block.
The first six come from the synthesizer's warnings. The last is detected in the source code, because Yosys accepts such blocks without any message.
A design with a critical warning still counts as synthesizable and continues to the later levels, because its code still becomes hardware. The warning is stored as part of its label, so such designs can be excluded.

\subsubsection{Post-synthesis metrics}
Each synthesizable design is mapped to the standard cells of the open Nangate 45\,nm library with ABC~\cite{brayton2010abc}, which gives its cell count, area, logic depth, and numbers of flip-flops and latches.
Its critical-path delay comes from static timing analysis with OpenSTA on a delay-oriented mapping of the same design, which inserts the buffers that the area-oriented mapping leaves out.

\paragraph{Empty circuits}
A design can be synthesizable and still map to no cell at all, because synthesis removes all logic that does not affect an output.
Such an \emph{empty circuit} arises in three ways, shown with real RTLCoder designs in Figure~\ref{fig:empty}.
An output can be undefined, for example because the register that drives it is clocked by a signal that is not a port (Figure~\ref{fig:empty}a).
An output can be constant, for example because the logic that computes it gives the same value for every input; the AND gate of Figure~\ref{fig:empty}b computes $\overline{(\bar{a} \oplus b) \oplus (a \oplus \bar{b})}$, and since $\bar{a} \oplus b = a \oplus \bar{b}$, its output is always 1.
Finally, every output can be wired directly to an input, as in the concatenation of Figure~\ref{fig:empty}c.
% Real RTLCoder designs that synthesize to an empty circuit (zero cells), one per kind (paper/stats/csv/D4_zero_cell_kinds.csv):
% id 17336 (delay_gate_4stage, undefined output), id 10424 (and_gate, constant output), id 1765 (concat_module, wiring only).
% Code from RTLCoder; blank lines removed, port lists joined, long lines wrapped; in 17336 the wire delayed_A (= delay_reg[3]) is inlined; the comment is added.
\begin{figure}[tp]
\centering
\scriptsize
\noindent
\begin{tcolorbox}[exbox, fontupper=\scriptsize, colback=red!3, colframe=red!60!black, width=0.32\linewidth, title={(a) Undefined output}, equal height group=empty]
\begin{Verbatim}[fontsize=\fontsize{6.4}{7.6}\selectfont]
module delay_gate_4stage (
    input A, output X);
  reg [3:0] delay_reg;
  // clk is not a port
  always @(posedge clk)
    delay_reg <=
      {delay_reg[2:0], A};
  assign X = delay_reg[3];
endmodule
\end{Verbatim}
\tcblower
\raggedright
\texttt{clk} is never driven, so the register is never clocked, \texttt{X} is undefined, and synthesis removes the register.
\end{tcolorbox}\hfill
\begin{tcolorbox}[exbox, fontupper=\scriptsize, colback=red!3, colframe=red!60!black, width=0.32\linewidth, title={(b) Constant output}, equal height group=empty]
\begin{Verbatim}[fontsize=\fontsize{6.4}{7.6}\selectfont]
module and_gate(input a,
    input b, output wire out);
  wire not_a, not_b,
       temp1, temp2;
  assign not_a = ~a;
  assign not_b = ~b;
  assign temp1 = not_a ^ b;
  assign temp2 = a ^ not_b;
  assign out = ~(temp1 ^ temp2);
endmodule
\end{Verbatim}
\tcblower
\raggedright
\texttt{temp1} and \texttt{temp2} are always equal, so \texttt{out} is 1 for every input instead of \texttt{a \& b}.
\end{tcolorbox}\hfill
\begin{tcolorbox}[exbox, fontupper=\scriptsize, colback=black!3, colframe=black!35, width=0.32\linewidth, title={(c) Wiring only}, equal height group=empty]
\begin{Verbatim}[fontsize=\fontsize{6.4}{7.6}\selectfont]
module concat_module(
    input wire [7:0] in_hi,
    input wire [7:0] in_lo,
    output wire [15:0] out);
  assign out = {in_hi, in_lo};
endmodule
\end{Verbatim}
\tcblower
\raggedright
Correct, but every output bit is wired to an input bit, so no gate is needed.
\end{tcolorbox}
\caption{The three ways in which a synthesizable RTLCoder design becomes an empty circuit with zero cells. The first two are wrong designs that synthesis accepts without an error; the third is correct but contains no logic.}
\label{fig:empty}
\end{figure}

A design is \emph{effectively synthesizable} only if its netlist contains at least one cell, and only such designs continue to Level~2; the cell count of every design is released, so a user can still select the wiring-only designs.

\subsubsection{Results}
\begin{figure}[t]
  \centering
  \includegraphics[width=\linewidth]{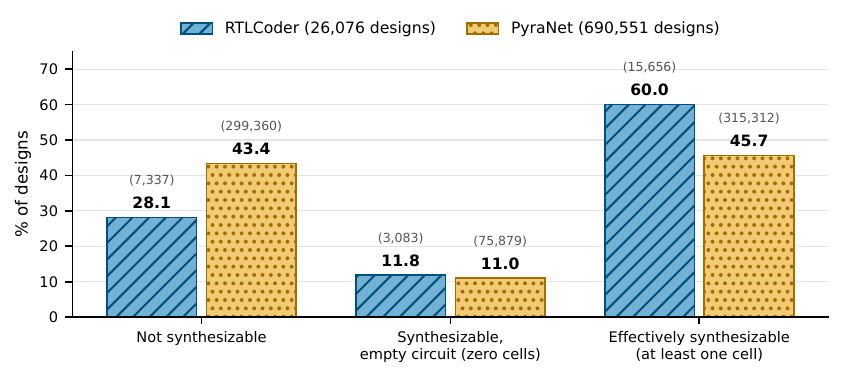}
  \caption{Level-1 outcome for each dataset, as a share of all designs (numbers of designs in parentheses). An empty circuit is synthesizable but maps to no standard cell.}
  \label{fig:level1}
\end{figure}

Figure~\ref{fig:level1} shows the outcome.
In RTLCoder, 71.9\% of the designs are synthesizable, but only 60.0\% are effectively synthesizable (15,656 designs), because 11.8\% of all designs synthesize to an empty circuit.
Re-synthesizing these empty circuits shows why. In 42\% of them, at least one output is never driven or undefined, in 30\% all outputs are constant, and in 19\% the outputs are only wired to inputs or constants, as in a byte reversal; only 2\% have no output at all, and the rest have other or mixed causes. The wired-only designs are correct hardware that needs no gates, so a user who wants them can select them by this label.
PyraNet is worse on both counts, with 56.6\% of its designs synthesizable and 45.7\% effectively synthesizable.
PyraNet's own quality rating of each design, given by an LLM, is almost unrelated to whether the design synthesizes (Spearman rank correlation 0.08), so an LLM rating is not a substitute for a synthesis check.
A pass label alone therefore overstates how much of a dataset is hardware, so cell count is released with every pass.

Many designs that synthesize are still probably defective.
Of the 14,385 effectively synthesizable RTLCoder designs processed at Levels~2--5, 16.3\% have at least one critical warning. The most common are conflicting drivers (7.1\%), undriven wires (3.6\%) and latches (3.1\%), followed by combinational loops (2.7\%), implicit declarations (2.6\%), out-of-range bit selects (2.5\%) and \texttt{initial} blocks (1.7\%).
In addition, 14.8\% have output bits that stay constant after synthesis, such as a flag that can never be raised.
Designs with a critical warning still produce hardware, so they continue to the later levels with the warning as part of their label, and 11.8\% of the designs accepted at Level~3 carry one.

\begin{keypoint}
\textbf{Key finding:} A large portion of public RTL training data is not synthesizable. 28\% of RTLCoder and 43\% of PyraNet designs fail synthesis, and 16\% and 19\% of the synthesizable ones produce no circuit. A model trained on such data without these labels learns from code that cannot generate RTL that can become hardware as if it were correct. Synthesizability should therefore be the first label of an RTL dataset, because it keeps non-hardware out of the positive examples and, together with the failure category, turns the failing designs into negative examples for contrastive learning and preference optimization.
\end{keypoint}

%% ------------------------------------------------------------
\subsection{Level 2: Specification quality}
\label{sec:level2}
Level~2 measures how well the original instruction specifies its design.
Figure~\ref{fig:ex-level2} shows the grades of a real instruction.
% Real records for Levels 1-3. Example design: RTLCoder id 53 (digital_system, batch 9), combinational; it matches no
% VerilogEval or RTLLM problem (Rouge-L below the exclusion thresholds). Level 1: ids 53 (passes, no critical warning),
% 16711 (altiobuf_in, fails), 1428 (power_good, passes with a critical warning). Level 2: grades of id 53 and its gaps,
% restated in plain words with an example input (the stored gaps are scores.*.gaps); rebuild results are orig and
% orig_iface (original instruction plus the exact port list), whose counterexample is from attempt 3 (all five attempts
% treat a,b,c,d as a 4-bit number). Level 3: its refined specification (accepted in the first round). Texts verbatim from
% spec_refinement/full_pilot/runs/portion09/results.jsonl and portions/p09/designs.jsonl; the instruction's opening
% sentence ("Please act as a professional Verilog designer.") is replaced by [...]; the four input declarations of
% the code are joined on one line, and "end" / "else begin" on one line; only highlighting is added.
% Level-1 numbers: paper/stats/cache/labels.parquet and cross_corpus_taxonomy/runs/rtlcoder_timing/timing.jsonl.
\begin{figure}[tp]
\centering
\scriptsize
\noindent\textbf{Example design} (RTLCoder design \texttt{digital\_system}, combinational)\par\smallskip
\noindent
\begin{tcolorbox}[exbox, fontupper=\scriptsize, colback=black!3, colframe=black!35, width=0.53\linewidth, title={Original instruction}, equal height group=lva]
[...] Create a module that implements a digital system with four inputs (a,b,c,d) that generates a logic-1 when {\sethlcolor{hlbad}\hl{the input is a multiple of 3}}, and a logic-0 otherwise. You should design the module using a combinational circuit and {\sethlcolor{hlbad}\hl{optimize the output expression to minimize the number of gates}}.
\end{tcolorbox}\hfill
\begin{tcolorbox}[exbox, fontupper=\scriptsize, colback=black!3, colframe=black!35, width=0.455\linewidth, title={Original code}, equal height group=lva]
\begin{Verbatim}[fontsize=\fontsize{6.4}{7.6}\selectfont]
module digital_system (
  input a, input b, input c, input d,
  output reg out
);
  always @* begin
    if ((a+b+c+d)%3 == 0) begin
      out <= 1;
    end else begin
      out <= 0;
    end
  end
endmodule
\end{Verbatim}
\end{tcolorbox}

\smallskip
\noindent\textbf{Level 2: grades of the original instruction} (grade, and the gaps that the scorer reports, in plain words)\par\smallskip
\noindent
\begin{tcolorbox}[exbox, fontupper=\scriptsize, colback=orange!3, colframe=orange!60!black, width=\linewidth, title={Scorer output}]
\begin{tabular}{@{}l c p{0.76\linewidth}@{}}
Consistency & 5 & No gap: what the instruction states is true of the code. \\
Unambiguity & 3 & ``The input is a multiple of 3'' does not say whether a, b, c, d form one 4-bit number or are four bits to be added: the two readings disagree, for example on \texttt{1100} (as a number, 12 is a multiple of 3; as a sum, $1{+}1{+}0{+}0=2$ is not) and on \texttt{1110} (as a number, 14 is not; as a sum, $1{+}1{+}1{+}0=3$ is). \\
Completeness & 2 & The module name, the output name and the port directions are missing. \\
Verifiability & 3 & ``Minimize the number of gates'' gives no gate count or limit, so no test can check it.
\end{tabular}\par\smallskip
\tcblower
Rebuilt from the original instruction: \textbf{0 of 5} attempts verified equivalent (all five use other port names). With the exact port list added, still \textbf{0 of 5}: all five read a, b, c, d as one 4-bit number, so for $a{=}1, b{=}1, c{=}0, d{=}0$ they output 1 and the code outputs 0.
\end{tcolorbox}
\caption{A real record of Level~2. The scorer grades the original instruction of \texttt{digital\_system} on the four qualities and reports what is missing, shown here in plain words ({\sethlcolor{hlbad}\hl{red}}: phrases that leave the behavior open). No circuit rebuilt from the instruction is equivalent to the code, even when the exact port list is added.}
\label{fig:ex-level2}
\end{figure}

\subsubsection{Four qualities of a specification}
A synthesizable design is useful for training only if its instruction describes it well enough that a designer could build it.
A specification can fail its reader in four different ways. It can state something false, state something that can be read in more than one way, leave something out, or state something that cannot be checked.
Each of these failures is measured by one quality, called \emph{consistency}, \emph{unambiguity}, \emph{completeness} and \emph{verifiability}.

The names of these qualities come from the IEEE recommended practice for software requirements specifications~\cite{ieee830}, of which Level~2 keeps only the characteristics that decide which circuit is built. It differs from the standard study~\cite{ieee830} in three ways.
(i)~Each quality is defined by what can be observed at the ports of the circuit.
(ii)~Consistency is checked against the paired code instead of within the text, so it covers both the \emph{correct} and the \emph{consistent} characteristics of the standard.
(iii)~The grades are tested against a tool result, which is whether the circuit can be rebuilt from the specification by the round trip of Level~3 (Section~\ref{sec:rubric-valid}).

The four qualities describe different defects, and a specification can be good in three of them and still lead to the wrong circuit because of the fourth (Section~\ref{sec:metric-overlap} measures how often this happens).
Together they describe what Level~3 tests directly, because a specification that is consistent, unambiguous, complete, and verifiable allows exactly one behavior at the ports, so any correct implementation of it is functionally equivalent to the original design.
Table~\ref{tab:rubric} gives the grades of each quality.

\paragraph{Consistency: every statement is true of the design.}
A specification can be clear, complete and testable and still describe a different circuit.
Consider the RTLCoder design \texttt{and4b}, whose instruction asks for the AND of its four inputs \texttt{A\_N}, \texttt{B}, \texttt{C} and \texttt{D}, while its code computes \texttt{X = \textasciitilde(A\_N | B | C | D)}, which is a NOR.
Each sentence of the instruction is precise, and a test for it is easy to write, but the pair teaches a model that this code computes an AND.
Such errors are common when the instruction and the code were written separately, for example when an LLM wrote the instruction for existing code.
None of the other three qualities can detect this, because they judge the specification alone; consistency is the only quality that compares the specification with the design.

\paragraph{Unambiguity: every statement has only one reading.}
A statement can be true of the design and still allow other designs.
Consider ``Output the average of the two 8-bit inputs \texttt{a} and \texttt{b}, rounded down.''
If the inputs are unsigned, the average of $\texttt{a}=\texttt{8'hFF}$ (255) and $\texttt{b}=\texttt{8'h01}$ (1) is 128; if they are signed, \texttt{a} is $-1$ and the average is 0.
Both circuits are reasonable readings, and a model trained on this pair learns that either one is correct.
Typical sources of ambiguity in Verilog instructions are the signedness of values, the clock edge (``on each clock''), the reset level (``an active reset'') and the order of priority between control signals.

\paragraph{Completeness: every behavior visible at the ports is stated.}
A specification can be true and unambiguous in everything it says and still be silent about part of the design.
Consider ``Design a debouncer for a push button.''
Nothing in this instruction is false or ambiguous, but the reader must guess how many clock cycles the input must stay stable, whether the output is active high or low, its value after reset, and what it shows while the input still bounces.
Each guess is a separate chance to build a different circuit.
Missing information is the most common defect in RTLCoder, where 73\% of the instructions receive a completeness grade of 3 or less, more than for any other quality.
The instruction of Figure~\ref{fig:ex-level2} is a typical case, since it never names the module or its output and never says which ports are inputs and which are outputs.

\paragraph{Verifiability: every requirement can be checked at the ports.}
A requirement is useful only if one can decide whether a design meets it.
Consider ``The debouncer should ignore glitches reliably and respond quickly.''
No test can decide what ``reliably'' or ``quickly'' means, so two very different designs satisfy this specification equally well.
A verifiable version states an observable condition, for example ``\texttt{out} takes the value of \texttt{btn} only after \texttt{btn} has kept that value for 16 consecutive rising edges of \texttt{clk}.''
Verifiability matters more for hardware than for most software, because the acceptance test of Level~3 compares designs only through their ports and clock cycles; a requirement that cannot be expressed in these terms cannot be confirmed by any tool.

\subsubsection{Why the scorer uses a rubric}
LLMs used as judges of text are known to depend on the order and length of the texts they compare, to favor their own text~\cite{zheng2023judging,wang2024fair,panickssery2024selfpref}, and to state facts that are not in their input~\cite{ji2023hallucination}. A \emph{rubric}, which states for each grade what a specification must contain, makes LLM grades agree more closely with human experts and easier to reproduce~\cite{liu2023geval,kim2024prometheus,ye2024flask,hashemi2024llmrubric}, and rubric scores now serve as training rewards where no exact answer exists~\cite{gunjal2025rubrics}.
The scorer of Level~2 uses such a rubric and adds four safeguards.
First, it judges the specification against what the design actually does, using facts extracted by Yosys and a behavior description that is checked against these facts (Section~\ref{sec:scorer-knows}), rather than its own reading of the code.
Second, every grade is defined by an example specification of a reference design, an 8-bit counter, and all grades use this same design, so the scale is fixed in advance (Table~\ref{tab:rubric}).
Third, the scorer must quote the specification and the behavior description as evidence for each grade, which makes unsupported grades visible.
Fourth, each defect counts against exactly one quality, and when the scorer hesitates between two grades it must give the lower one.
The grades are not taken on trust. Sections~\ref{sec:rubric-valid} and~\ref{sec:crossjudge} test whether they predict the outcome of a tool-based check and whether a second LLM from another model family gives the same grades.

\subsubsection{What the scorer knows about the design}
\label{sec:scorer-knows}
% Real record: RTLCoder design bin_to_gray (id 23012, portion 08). Code from the design with the begin/end
% of the always block removed and indentation normalized (same behavior); facts verbatim from the request logged in spec_refinement/full_pilot/runs/portion08/api_calls/23012.jsonl.
\begin{figure}[tp]
\centering
\scriptsize
\noindent
\begin{tcolorbox}[exbox, fontupper=\scriptsize, colback=black!3, colframe=black!35, width=0.42\linewidth, title={Verilog code}, equal height group=facts]
\begin{Verbatim}[fontsize=\fontsize{6.4}{7.6}\selectfont]
module bin_to_gray(
  input clk,
  input rst,
  input [3:0] BIN,
  output reg [3:0] GRAY
);
  always @(posedge clk)
    if (rst)
      GRAY <= 4'b0000;
    else
      GRAY <= BIN ^ (BIN >> 1);
endmodule
\end{Verbatim}
\end{tcolorbox}\hfill
\begin{tcolorbox}[exbox, fontupper=\scriptsize, colback=green!4, colframe=green!45!black, width=0.55\linewidth, title={Facts extracted by Yosys}, equal height group=facts]
\begin{Verbatim}[fontsize=\fontsize{6.4}{7.6}\selectfont]
Top module: bin_to_gray
Ports:
  input  clk     1 bit
  input  rst     1 bit
  input  BIN     4 bits
  output GRAY    4 bits
Clocks: clk (rising edge)
Resets: rst (synchronous, active high)
Storage: 4 flip-flop bits, 0 latch bits, 0 memory bits
Parameters: none
Outputs:
  GRAY: registered, reset value 0
\end{Verbatim}
\end{tcolorbox}
\caption{The facts that Yosys extracts from an RTLCoder design (right), shown with its code (left). The facts give the ports, clock, reset, storage and the kind and reset value of each output, but not what the output computes; that part comes from the behavior description.}
\label{fig:facts}
\end{figure}

To grade a specification, the scorer must know what the design actually does. An LLM reading the code alone can misread it, so the scorer receives two descriptions of the design that complement each other.
(i)~\emph{Facts extracted by Yosys} (Figure~\ref{fig:facts}). They are exact but describe only the structure of the design, such as its ports with their direction, width and sign, its parameters, its clocks and their edges, its resets with their type and active level, its memory size, and, for each output, whether it is stored in a register and its reset value. They do not say what an output computes.
(ii)~\emph{A behavior description} that the refiner writes from the code and the facts (the \emph{Detective} in Figure~\ref{fig:framework}). It fills this gap. For every output, it states the conditions that change it in priority order, its delay in clock cycles, and any behavior that the code leaves undefined (Figure~\ref{fig:prompts}a). It describes only what the code does and ignores the comments and the instruction.
Because the behavior description is written by an LLM, it is checked against the exact facts. If the two disagree, for example on the width of a port or the active level of a reset, the LLM has misread the code, and the design is labeled \emph{misunderstood} and goes no further; this happened for 57 of the 14,385 designs.

\subsubsection{Grades and prompt}
The refiner, acting as the scorer (the \emph{Grader} in Figure~\ref{fig:framework}), gives each quality a grade from 1 (poor) to 5 (excellent).
Table~\ref{tab:rubric} lists the grades, and Figure~\ref{fig:prompt-scoring} shows the scoring prompt.

% Scoring prompt of Level 2, condensed: score_system.txt with the grade definitions replaced by a pointer to Table tab:rubric.
% The full verbatim prompt is in Appendix A (sections/A_full_prompts.tex).
\begin{figure}[t]
\begin{tcolorbox}[enhanced, arc=3pt, boxrule=1pt, colback=navy!3, colframe=navy, left=4pt, right=4pt, top=2pt, bottom=2pt,
  title={\sffamily\bfseries\small Scoring prompt (system message, condensed)}, coltitle=white, colbacktitle=navy]
{\ttfamily\scriptsize\raggedright\frenchspacing\setlength{\parindent}{0pt}
\mbox{}You are a senior digital-{}design engineer reviewing a hardware specification.\\
\mbox{}You will be given a DESCRIPTION of a design, a verified BEHAVIOR SHEET of the\\
\mbox{}design (ground truth), and tool facts. Score the DESCRIPTION on four metrics\\
\mbox{}using the rubric. Do not rewrite it.\\
\mbox{}\\
\mbox{}================ RUBRIC ================\\
\mbox{}Each defect is charged to ONE metric:\\
\mbox{}-{} stated and wrong, or self-{}contradictory\hphantom{xxxxxx}-{}\textgreater{} CONSISTENCY\\
\mbox{}-{} stated but has several readings\hphantom{xxxxxxxxxxxxxx}-{}\textgreater{} UNAMBIGUITY\\
\mbox{}-{} not stated at all\hphantom{xxxxxxxxxxxxxxxxxxxxxxxxxxxx}-{}\textgreater{} COMPLETENESS\\
\mbox{}-{} stated but not checkable at the ports\hphantom{xxxxxxxx}-{}\textgreater{} VERIFIABILITY\\
\mbox{}Score against the behavior sheet, not your own reading of the code.\\
\mbox{}If torn between two scores, give the lower. For purely combinational\\
\mbox{}designs, clock/reset/latency items are N/A and count as satisfied.\\
\mbox{}\\
\mbox{}Reference example (ground truth): counter8 -{} 8-{}bit up-{}counter; rising edge\\
\mbox{}of clk; synchronous active-{}high rst clears count to 0 and has priority over\\
\mbox{}en; else if en, count increments and wraps 255 -{}\textgreater{} 0; count is registered;\\
\mbox{}tc (combinational) = 1 exactly when count == 255 and en == 1.\\
\mbox{}\\
\mbox{}[Grade definitions 5 to 1 for each metric, each with an example\\
\mbox{}\hphantom{x}of the reference design, as in the rubric table.]\\
\mbox{}========================================\\
\mbox{}\\
\mbox{}Return only JSON:\\
\mbox{}\{\\
\mbox{}\hphantom{xx}\textquotedbl{}consistency\textquotedbl{}:\hphantom{xxx}\{\textquotedbl{}score\textquotedbl{}: 1-{}5, \textquotedbl{}evidence\textquotedbl{}: \textquotedbl{}quote the description; cite the behavior-{}sheet entry\textquotedbl{}\},\\
\mbox{}\hphantom{xx}\textquotedbl{}unambiguity\textquotedbl{}:\hphantom{xxx}\{\textquotedbl{}score\textquotedbl{}: 1-{}5, \textquotedbl{}evidence\textquotedbl{}: \textquotedbl{}...\textquotedbl{}\},\\
\mbox{}\hphantom{xx}\textquotedbl{}completeness\textquotedbl{}:\hphantom{xx}\{\textquotedbl{}score\textquotedbl{}: 1-{}5, \textquotedbl{}evidence\textquotedbl{}: \textquotedbl{}...\textquotedbl{}\},\\
\mbox{}\hphantom{xx}\textquotedbl{}verifiability\textquotedbl{}: \{\textquotedbl{}score\textquotedbl{}: 1-{}5, \textquotedbl{}evidence\textquotedbl{}: \textquotedbl{}...\textquotedbl{}\}\\
\mbox{}\}\par}
\end{tcolorbox}
\caption{The system prompt that grades a text on the four qualities, with the grade definitions of Table~\ref{tab:rubric} left out. The same prompt is used for the original instructions, the refined specifications, and the second scorer. The user message contains the text to grade, the Yosys facts and the checked behavior description.}
\label{fig:prompt-scoring}
\end{figure}

% Grade descriptions of the Level-2 rubric, copied from spec_refinement/prompts/score_system.txt (verbatim content,
% shortened wording only where marked by the column layout). The full prompt is in Appendix app:prompt.
\begin{table}[t]
\centering
\footnotesize
\caption{Grades of the four specification qualities, as given to the scoring LLM. The examples describe the same reference design: an 8-bit up-counter with a synchronous active-high reset that has priority over the enable, wrap-around from 255 to 0, a registered output \texttt{count}, and a combinational output \texttt{tc} that is 1 exactly when \texttt{count} is 255 and \texttt{en} is 1.}
\label{tab:rubric}
\renewcommand{\arraystretch}{1.15}
\setlength{\tabcolsep}{4pt}
\begin{tabularx}{\linewidth}{@{}c>{\raggedright\arraybackslash}X>{\raggedright\arraybackslash}X>{\raggedright\arraybackslash}X>{\raggedright\arraybackslash}X@{}}
\toprule
Grade & \textbf{Consistency}: every statement is true of the design & \textbf{Unambiguity}: every statement has one reading & \textbf{Completeness}: every port-visible behavior is stated & \textbf{Verifiability}: every requirement can be checked at the ports \\
\midrule
5 & Every stated behavior matches. & Nothing is open. & An equivalent design can be built from the text alone. & Every requirement maps to a check on named ports. \\
4 & One secondary detail is wrong: \emph{``tc is a registered output.''} & One minor item is open: \emph{``on each clock''} (which edge?) & One minor item is missing: priority of reset over enable not stated. & One requirement has no time or condition: \emph{``tc indicates the last count.''} \\
3 & Core correct, several details wrong: \emph{``asynchronous reset \dots\ tc high at 254.''} & One or two important items are open: \emph{``active reset''}; \emph{``tc signals overflow.''} & Ports complete, several behaviors missing: \texttt{tc} never defined, wrap-around not stated. & Several vague requirements: \emph{``updates as needed''}; \emph{``at the right time.''} \\
2 & Core partly wrong or self-contradictory: \emph{``8-bit counter that counts 0 to 999, then saturates.''} & Most behavior is open: \emph{``reset clears it; counts when enabled; tc marks the end.''} & Ports incomplete, core partly given: \emph{``8-bit counter with clock and reset; output count.''} & Mostly qualitative: \emph{``works correctly; handles overflow reliably.''} \\
1 & A different function: \emph{``a 4-bit down-counter with load.''} & The core function is open: \emph{``a counter with reset and enable that signals when done.''} & Almost nothing is stated: \emph{``Design a counter.''} & Entirely qualitative: \emph{``a fast, robust, low-power counter.''} \\
\bottomrule
\end{tabularx}
\par\smallskip\footnotesize Each defect is charged to one quality: stated and wrong (consistency), stated with several readings (unambiguity), not stated (completeness), stated but not checkable at the ports (verifiability). For combinational designs, clock, reset and latency items count as satisfied. If the scorer is torn between two grades, it gives the lower one. A higher grade is better (\hib).
\end{table}

\subsubsection{Do the grades predict anything?}
\label{sec:rubric-valid}
\begin{figure}[t]
  \centering
  \begin{minipage}[b]{0.64\linewidth}\centering
    \includegraphics[width=\linewidth]{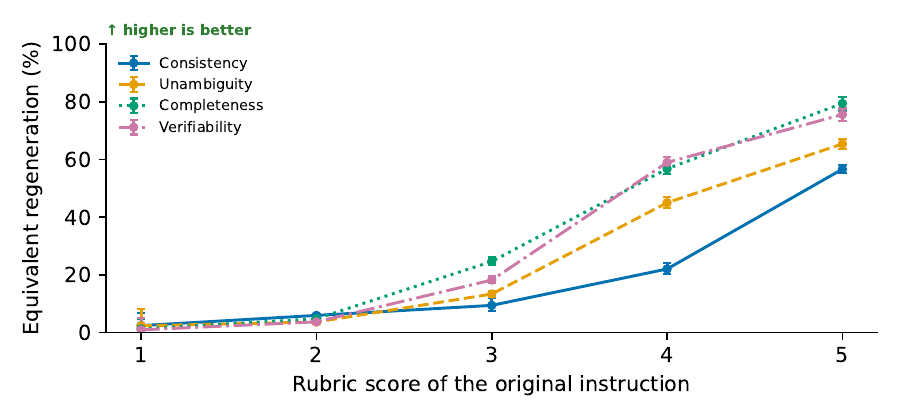}\\[-1pt]
    {\small (a)}
  \end{minipage}\hfill
  \begin{minipage}[b]{0.33\linewidth}\centering
    \includegraphics[width=\linewidth]{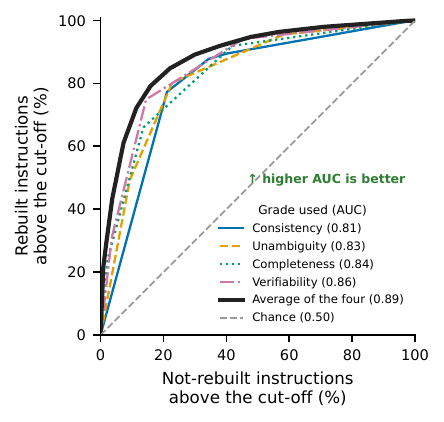}\\[-1pt]
    {\small (b)}
  \end{minipage}
  \caption{(a)~Share of designs that another LLM rebuilds from the original instruction, by the instruction's grade on each quality. (b)~ROC curve of each grade and of their average as a predictor of whether the circuit is rebuilt, with the AUC in parentheses; the dashed diagonal is chance.}
  \label{fig:rubric-validity}
\end{figure}

If the four qualities capture what makes a specification usable, a specification with high grades should let another model rebuild the circuit, and one with low grades should not.
The round trip of Level~3, described next, tests this when it is applied to the original instructions.
Figure~\ref{fig:rubric-validity}a shows that the grades predict the outcome well.
This test uses the 14,292 designs that have both grades and a round-trip result, that is, the 14,385 designs without the 57 misunderstood ones and the 36 whose refined specification never passes the program checks of Level~3 (Section~\ref{sec:level3}). Of these, those whose instruction has an average grade below 3 are rebuilt in less than 5\% of cases, and those with an average of 4 or more in 69.3\% of cases.
The average of the four grades separates the rebuilt from the not-rebuilt instructions with an area under the ROC curve (AUC) of 0.89, where 0.5 is chance and 1.0 is perfect (Figure~\ref{fig:rubric-validity}b).
Taken alone, each quality also predicts the outcome, and verifiability and completeness do so best, with an AUC of 0.86 and 0.84, against 0.83 for unambiguity and 0.81 for consistency (Figure~\ref{fig:rubric-validity}b). Their rebuild rates rise most steeply with the grade (Figure~\ref{fig:rubric-validity}a); for example, an instruction with a completeness grade of 2 is rebuilt in 4.8\% of cases, and one with a grade of 5 in 79.4\% of cases.

\subsubsection{Are the four qualities distinct?}
\label{sec:metric-overlap}
\begin{figure}[t]
  \centering
  \includegraphics[width=\linewidth]{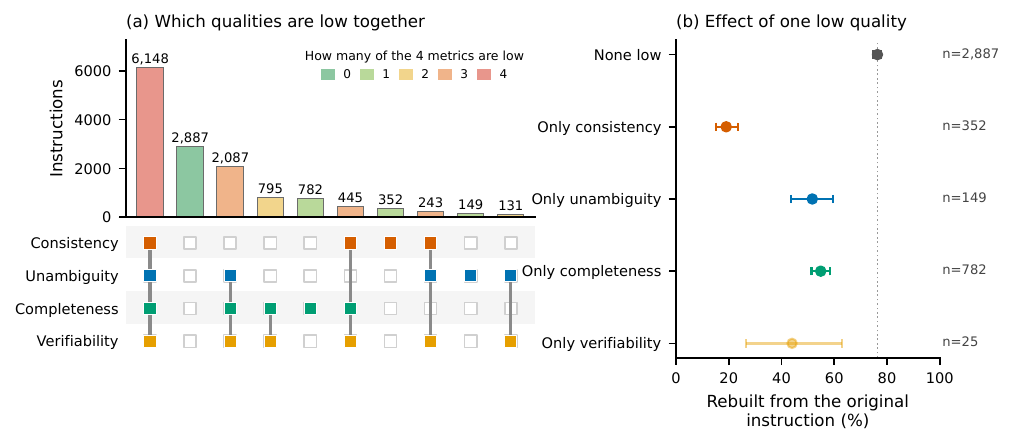}
  \caption{Overlap of the four qualities on the 14,292 graded original instructions; a quality is low if its grade is 3 or less. (a)~Number of instructions for each combination of low qualities, for the ten most frequent combinations; a colored square marks a low quality, and the bar color gives how many of the four qualities are low. (b)~Share of instructions rebuilt from the original instruction when no quality is low and when exactly one quality is low.}
  \label{fig:metric-overlap}
\end{figure}

\begin{figure}[t]
  \centering
  \includegraphics[width=\linewidth]{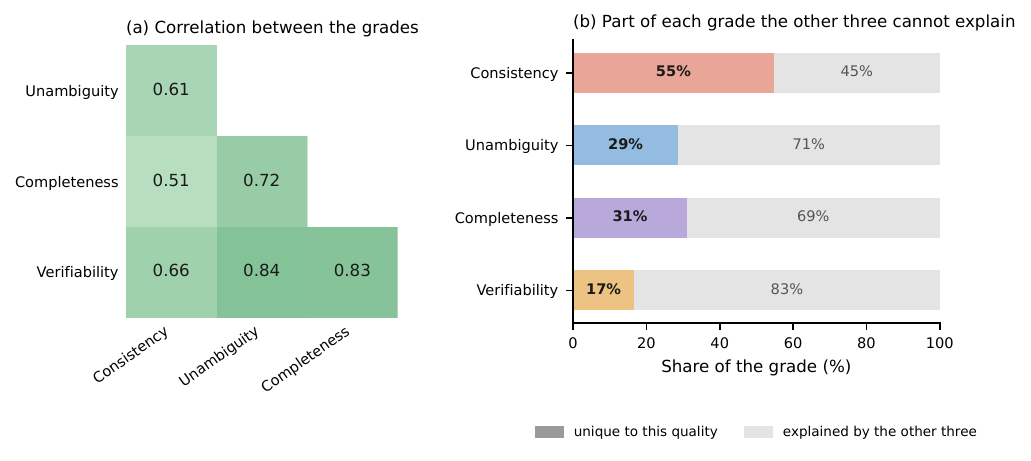}
  \caption{The four qualities are related but not redundant (14,292 graded original instructions). (a) Spearman correlation between the grades of each pair of qualities. (b) For each quality, the part of its grade that the other three grades explain (gray; $R^2$ of a linear regression on the ranks of the other three) and the part that they cannot explain (color). A quality that duplicated the others would have almost no colored part.}
  \label{fig:metric-corr}
\end{figure}

The grades of the four qualities are correlated (Spearman rank correlation 0.51--0.84; Figure~\ref{fig:metric-corr}a), because a poor instruction is often poor in several ways, but no pair is close to 1, and 17--55\% of each grade cannot be explained by the other three (Figure~\ref{fig:metric-corr}b).
Consistency is the most distinct, since it is the only quality that compares the instruction with the code.
Verifiability shares the most with unambiguity and completeness, as expected, because a requirement that is missing or has two readings cannot be checked either.
Its own role is to catch requirements that are stated clearly but that no test at the ports can check, such as ``make it fast'' or ``use as few gates as possible''.
Such a requirement does not describe what the circuit does, so the refined specification must drop it or turn it into a checkable one, as with the gate-count request of Figure~\ref{fig:ex-level2}.
Correlation alone cannot show whether one quality is only a part of another, so Figure~\ref{fig:metric-overlap}a counts, for every instruction, which qualities are low (a grade of 3 or less).
In 43\% of the instructions all four are low, and in 20\% none is.
Consistency, completeness and unambiguity are each the only low quality in 352, 782 and 149 instructions, so each carries information about defects that the other three do not carry.
Verifiability is rarely the only low quality (25 instructions), because an uncheckable requirement usually comes with missing or ambiguous ones.
Each of these three qualities also matters on its own (Figure~\ref{fig:metric-overlap}b). When only one of them is low, the share of instructions that another LLM rebuilds falls from 76.4\% (no low quality) to 54.9\% for completeness, 51.7\% for unambiguity and 19.0\% for consistency.

\subsubsection{Do the grades depend on the scoring model?}
\label{sec:crossjudge}
\begin{figure}[t]
  \centering
  \begin{minipage}[b]{0.70\linewidth}\centering
    \includegraphics[height=4cm]{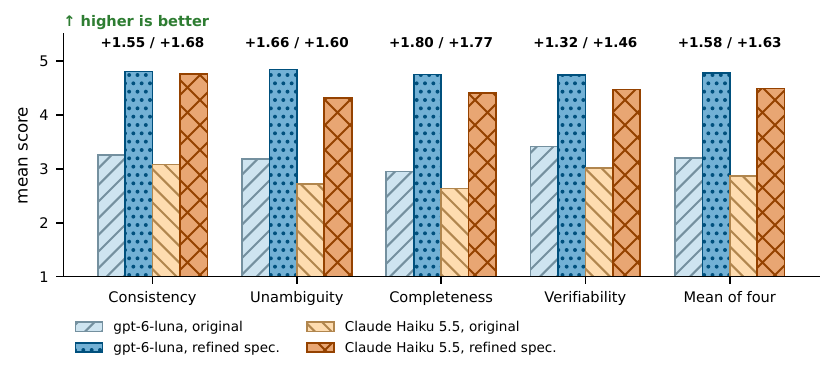}\\[-1pt]
    {\small (a)}
  \end{minipage}\hfill
  \begin{minipage}[b]{0.28\linewidth}\centering
    \includegraphics[height=4cm]{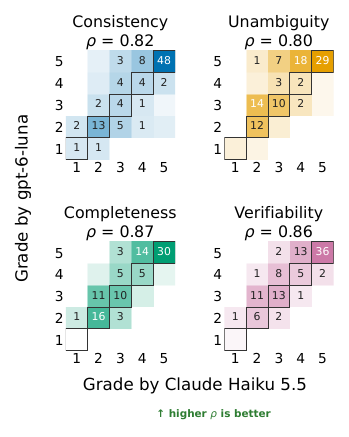}\\[-1pt]
    {\small (b)}
  \end{minipage}
  \caption{Grades given by gpt-6-luna and Claude Haiku~5.5 to the same 297 designs. (a)~Average grades; above the bars, the improvement from the original instruction to the refined specification (gpt-6-luna / Haiku). (b)~Share (\%) of the 594 graded texts that received each pair of grades; outlined cells are exact agreement, and $\rho$ is the Spearman correlation.}
  \label{fig:crossjudge}
\end{figure}

The grades of Level~2 are given by gpt-6-luna, the same LLM that later rewrites the instructions.
A scorer might favor its own writing, or another model might grade differently.
To test this, a second LLM from another model family, Claude Haiku~5.5, graded 297 random designs with the same prompt, both the original instruction and the refined specification.
Figure~\ref{fig:crossjudge}a shows the result.
The two scorers agree closely. The Spearman rank correlation of their grades is 0.80--0.87 for the four qualities and 0.91 for the average, and for each quality, 91--97\% of their grades differ by at most one point (Figure~\ref{fig:crossjudge}b).
Haiku is somewhat stricter in absolute terms, but it finds the same improvement from the original instruction to the refined specification, $+1.63$ points on average, against $+1.58$ for gpt-6-luna.
The difference, $-0.05$, shows no preference of the rewriting model for its own specifications.
The grades therefore describe the specification, not the model that gives them.

\begin{keypoint}
\textbf{Key finding:} The grades of four qualities predict whether an instruction is precise enough to rebuild its circuit (ROC AUC 0.89), and two LLMs from different model families (OpenAI vs Anthropic) give nearly the same grades (Spearman 0.91), with no preference for their own specifications. The grades therefore describe the specification itself and can screen training instructions before any circuit is rebuilt.
\end{keypoint}

%% ------------------------------------------------------------
\subsection{Level 3: The refined specification determines the circuit}
\label{sec:level3}
Level~3 rewrites the original instruction until it determines its circuit.
Figure~\ref{fig:ex-level3} shows an accepted refined specification (a) and a refinement trajectory (b).
\begin{figure}[tp]
\centering
\scriptsize
\raggedright\noindent\textbf{(a) Accepted in the first round} (RTLCoder design \texttt{digital\_system}, combinational)\par\smallskip
\noindent
\begin{tcolorbox}[exbox, fontupper=\scriptsize, colback=black!3, colframe=black!35, width=0.53\linewidth, title={Original instruction (as in Figure~\ref{fig:ex-level2})}, equal height group=lva3]
[...] Create a module that implements a digital system with four inputs (a,b,c,d) that generates a logic-1 when {\sethlcolor{hlbad}\hl{the input is a multiple of 3}}, and a logic-0 otherwise. You should design the module using a combinational circuit and {\sethlcolor{hlbad}\hl{optimize the output expression to minimize the number of gates}}.
\end{tcolorbox}\hfill
\begin{tcolorbox}[exbox, fontupper=\scriptsize, colback=black!3, colframe=black!35, width=0.455\linewidth, title={Original code}, equal height group=lva3]
\begin{Verbatim}[fontsize=\fontsize{6.4}{7.6}\selectfont]
module digital_system (
  input a, input b, input c, input d,
  output reg out
);
  always @* begin
    if ((a+b+c+d)%3 == 0) begin
      out <= 1;
    end else begin
      out <= 0;
    end
  end
endmodule
\end{Verbatim}
\end{tcolorbox}

\smallskip
\noindent
\begin{tcolorbox}[exbox, fontupper=\scriptsize, colback=green!4, colframe=green!45!black, width=\linewidth, title={Refined specification (accepted)}]
Overview: digital\_system produces a logic-1 when {\sethlcolor{hlfix}\hl{the sum of its four input bits is divisible by 3}}, and a logic-0 otherwise.
Interface: {\sethlcolor{hlfix}\hl{Module digital\_system. Inputs: a, b, c, d, each 1-bit unsigned. Output: out, 1-bit unsigned.}}
Behavior: {\sethlcolor{hlfix}\hl{For every combination of a, b, c, and d, out is 1 exactly when the sum of their individual bit values is divisible by 3; otherwise out is 0.}} The result is combinational with no cycle latency.
\tcblower
Rebuilt: \textbf{5 of 5} attempts verified equivalent (a proof over all 16 inputs). The green parts answer the gaps of Level~2: the meaning of ``multiple of 3'' and the ports. The gate-count request, which no test can check, is not part of the refined specification.
\end{tcolorbox}

\medskip
\noindent\textbf{(b) Accepted after one revision} (RTLCoder design \texttt{adder}, a saturating adder, combinational)\par\smallskip
\noindent
\begin{tcolorbox}[exbox, fontupper=\scriptsize, colback=black!3, colframe=black!35, width=0.6\linewidth, title={Original instruction (shortened)}, equal height group=lvb]
You are tasked with designing a Verilog module that will take in two 8-bit inputs, A and B, and output a single 8-bit output, Y. The output Y is the result of adding A and B, but with a twist: if the sum of A and B is greater than 255, then the output Y should be set to 255. Your module should be named ``adder'' [...]
\end{tcolorbox}\hfill
\begin{tcolorbox}[exbox, fontupper=\scriptsize, colback=black!3, colframe=black!35, width=0.385\linewidth, title={Original code}, equal height group=lvb]
\begin{Verbatim}[fontsize=\fontsize{6.4}{7.6}\selectfont]
module adder(input [7:0] A, input [7:0] B,
             output reg [7:0] Y);
always @(*) begin
  if (A + B > 255) begin
    Y <= 255;
  end else begin
    Y <= A + B;
  end
end
endmodule
\end{Verbatim}
\end{tcolorbox}

\smallskip
\noindent
\begin{tcolorbox}[exbox, fontupper=\scriptsize, colback=red!3, colframe=red!60!black, width=0.36\linewidth, title={First refined specification (fails)}, equal height group=lv3]
[...] Behavior: {\sethlcolor{hlbad}\hl{Compute A + B as an 8-bit unsigned value, wrapping modulo 256.}} If this wrapped value is greater than 255, Y is 255; otherwise Y is the wrapped value. Thus, for every input pair, Y is the low 8 bits of A + B.
\tcblower
Rebuilt: \textbf{0 of 5} verified equivalent.
\end{tcolorbox}\hfill
\begin{tcolorbox}[exbox, fontupper=\scriptsize, colback=black!3, colframe=black!35, width=0.25\linewidth, title={Counterexample}, equal height group=lv3]
\centering
\begin{tabular}{@{}c c c c@{}}
A & B & \multicolumn{2}{c}{Y} \\
 & & orig. & rebuilt \\
255 & 128 & 255 & \textbf{\color{red!70!black}127} \\
\end{tabular}
\tcblower
\raggedright
In Verilog, \texttt{A + B > 255} is computed with more than 8 bits, so the sum does not wrap and the code saturates.
\end{tcolorbox}\hfill
\begin{tcolorbox}[exbox, fontupper=\scriptsize, colback=green!4, colframe=green!45!black, width=0.36\linewidth, title={Revised specification (accepted)}, equal height group=lv3]
[...] Behavior: {\sethlcolor{hlfix}\hl{Compare the full unsigned mathematical sum A + B against 255.}} If the sum is greater than 255, Y is 255; otherwise, Y is the low 8 bits of the sum.
\tcblower
Rebuilt: \textbf{5 of 5} verified equivalent. The trajectory stores both specifications and the counterexample.
\end{tcolorbox}
\caption{Real records of Level~3. (a)~The refined specification of \texttt{digital\_system} closes the gaps that Level~2 reported ({\sethlcolor{hlfix}\hl{green}}) and is accepted because every rebuild from it is proven equivalent to the code. (b)~A refinement trajectory. The first refined specification of a saturating adder wrongly states that the sum wraps at 8 bits; the counterexample $A=255$, $B=128$ shows that the code saturates instead, and the revised specification, which compares the full sum, is accepted.}
\label{fig:ex-level3}
\end{figure}

\subsubsection{The round trip}
The grades of Level~2 are given by an LLM, so they do not decide whether a specification is accepted.
Instead, each specification is tested directly with a \emph{round trip} (the Level~3 panel of Figure~\ref{fig:framework}).
A second LLM, the \emph{builder}, receives only the specification and writes the module five times (the \emph{Builders} in Figure~\ref{fig:framework}; Figure~\ref{fig:prompts}c).
An attempt that does not compile receives the Yosys error once and may correct itself.
A formal equivalence checker (the \emph{Equivalence check} box in Figure~\ref{fig:framework}) then compares each attempt with the original code, and the specification passes if at least one attempt is verified equivalent.
The builder (Claude Haiku~5.5) comes from a different model family than the refiner (gpt-6-luna), so it does not share the refiner's view of the code.

\subsubsection{The equivalence checker}
Two circuits are \emph{functionally equivalent} if they produce the same outputs for every sequence of inputs. A SAT solver applied to a \emph{miter}, a circuit that feeds the same inputs to both designs and signals any difference between their outputs, either confirms this or returns a \emph{counterexample}, an input sequence on which they differ.
Two designs that pass the check below are called \emph{verified equivalent}; for combinational designs this is a proof, and for sequential designs it is \emph{bounded} to a fixed number of clock cycles.
The checker works in four steps.
First, it compares the ports of the two designs, whose names, directions and widths must match.
Second, it builds a miter of the two designs with Yosys and asks the SAT solver whether any output can differ, starting from a state in which all registers are zero.
For combinational designs this is a proof over all inputs.
For sequential designs the check is bounded and covers the first 20 clock cycles after reset.
Third, because some differences appear only after many cycles, for example in a large counter, it simulates both designs with three random input sequences of 3,000 cycles each and with different reset patterns.
Fourth, it checks that both designs use the same clock edges, which the SAT model alone does not distinguish.
An attempt is verified equivalent only if all steps pass.

\subsubsection{Rewriting the instruction}
Because most original instructions fail this test, the refiner, shown as the \emph{Writer} in Figure~\ref{fig:framework}, rewrites every instruction.
It receives the original instruction, the Yosys facts and the checked behavior description, and writes a \emph{refined specification} with five fixed sections (overview, interface, behavior, reset and timing; Figure~\ref{fig:prompts}b).
The refined specification is \emph{functional}. It states what the outputs do, cycle by cycle, but not how the circuit computes them.
The refined specification is grounded in the code, because a training label must be true of the code it is paired with; otherwise the pair teaches a wrong mapping. Where the original instruction and the code disagree, the refined specification follows the code, and the record of the design is marked with a flag saying that the instruction contradicts the code, so that users can find or exclude such pairs (the flags are listed below).
The refined specification must describe behavior, not implementation, and must not contain code.
A program checks this. The refined specification must name every port, must not contain lines that look like Verilog, and must not share any sequence of 15 words with the code outside the port list.
A refined specification that fails is returned once for correction.
If the round trip then fails, the refiner receives the counterexample from the checker, cycle by cycle, and revises the refined specification (Figure~\ref{fig:prompts}d).
The prompt asks it to state the general rule that the counterexample exposes, not the counterexample itself.
This is repeated at most twice.

The code is never changed, even when it is probably wrong, for example when an output is stuck at a constant value; the refined specification then describes what the code actually does.
Such cases are recorded as \emph{flags} instead. The refiner flags an original instruction that contradicts the code or is incomplete and code that may have a bug, and Yosys flags structural problems such as undriven signals, several drivers on one signal, latches and constant outputs. The flags do not decide acceptance, but they let users find or exclude such designs.

% Condensed templates of the other prompts. Full verbatim prompts: Appendix (prompts/*.txt).
\newcommand{\ptbox}[3]{%
\begin{tcolorbox}[enhanced, nobeforeafter, box align=top, width=#1, arc=3pt, boxrule=0.9pt, colback=white,
  colframe=navy, left=3pt, right=3pt, top=1pt, bottom=2pt, fontupper=\scriptsize\sffamily,
  title={\sffamily\bfseries\footnotesize #2}, coltitle=black, colbacktitle=gold!35]
#3
\end{tcolorbox}}
\begin{figure}[t]
\centering
\ptbox{0.485\linewidth}{(a) Behavior description (refiner)}{%
\textbf{Input:} Verilog code; Yosys facts.\\
\textbf{Task:} ``Produce a precise behavior sheet: everything an observer at the module's ports could see, cycle by cycle.''\\
\textbf{Rules:} describe only what the code does; copy names, widths, clocks and resets from the Yosys facts; for every output give combinational or registered, latency, reset value, and every condition in priority order; list behavior the code leaves undefined; do not guess intent.\\
\textbf{Output:} JSON (clocking, reset, outputs, state machines, arithmetic). Checked against the Yosys facts.}\hfill
\ptbox{0.485\linewidth}{(b) Grades and refined specification (refiner)}{%
\textbf{Input:} original instruction; Yosys facts; behavior description (a).\\
\textbf{Task:} ``Part 1: score the original description on four metrics using the rubric. Part 2: write a refined specification that would score 5 on all four.''\\
\textbf{Rules:} the behavior description is the ground truth; exact interface; no code or pseudo-code; describe behavior, not implementation.\\
\textbf{Output:} grades with evidence; refined specification with sections Overview, Interface, Behavior, Reset, Timing; flags for contradiction, incompleteness and code defects.}

\medskip
\ptbox{0.485\linewidth}{(c) Builder (independent LLM, verbatim)}{%
``You are a hardware engineer. Implement the given specification as one synthesizable Verilog-2005 module (plus any submodules it needs). Reply with exactly one verilog code block and nothing else.''\\
\textbf{Input:} the text only. Five attempts; an attempt that does not compile gets one repair turn with the Yosys error.}\hfill
\ptbox{0.485\linewidth}{(d) Revision (refiner, verbatim)}{%
``A module generated from it by another engineer is NOT equivalent to the ground-truth design. Revise the specification so the failing behavior is stated unambiguously. [\dots] Describe the general rule that the counterexample exposes, not the counterexample itself. Change only what is needed.''\\
\textbf{Input:} behavior description; previous refined specification; counterexample, cycle by cycle after reset.}

\medskip
\ptbox{0.485\linewidth}{(e) Paraphrase (refiner)}{%
\textbf{Input:} accepted refined specification and its list of critical facts (each one rule, constant, priority or timing property).\\
\textbf{Task:} rewrite with one main change: lexical, syntactic or representational.\\
\textbf{Rules:} state every critical fact with the same meaning; add, remove, weaken or strengthen nothing; keep all names and widths; no code; do not copy sentences.\\
\textbf{Output:} text and, for every fact, the exact span that states it. Rejected if a fact is missing or more than 60\% of its four-word sequences come from the specification.}\hfill
\ptbox{0.485\linewidth}{(f) One-fact variant (refiner)}{%
\textbf{Input:} accepted refined specification, critical facts, reference code.\\
\textbf{Task:} ``Change exactly ONE critical fact so that the described behavior differs observably at the ports,'' for example an active level, a priority, a constant, a latency or a clock edge.\\
\textbf{Rules:} update every sentence that states the changed fact; keep all other sentences word for word; keep the interface; change the code minimally to match.\\
\textbf{Output:} changed refined specification and changed code. Accepted only if a counterexample shows that the changed code differs from the original.}
\caption{Templates of the other prompts used in Levels 2--5, with key instructions quoted.}
\label{fig:prompts}
\end{figure}

\subsubsection{Results}
\begin{figure}[t]
  \centering
  \includegraphics[width=\linewidth]{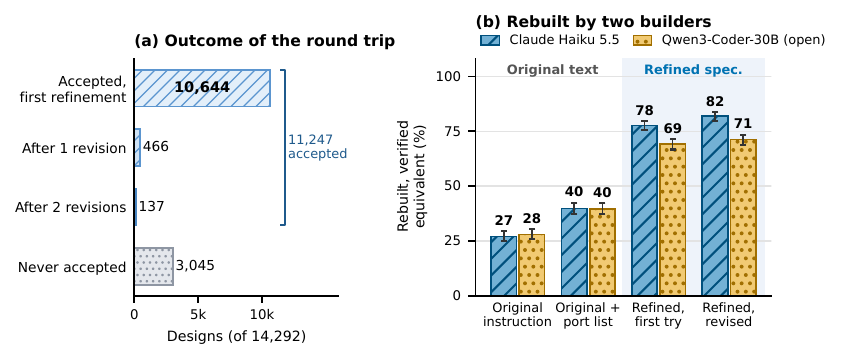}
  \caption{The round trip. (a) Outcome of the 14,292 designs that reach it; the 11,247 accepted designs give the refinement trajectories. (b) Share of designs that each builder rebuilds from each kind of specification, verified equivalent to the original code (1,460 randomly sampled designs; 95\% confidence intervals). From the original instruction to the refined specification after revision, Claude Haiku~5.5 gains 55 and the open Qwen3-Coder-30B 43 percentage points, so the gain is not tied to the builder that the pipeline uses.}
  \label{fig:roundtrip}
\end{figure}

\begin{figure}[t]
  \centering
  \includegraphics[width=\linewidth]{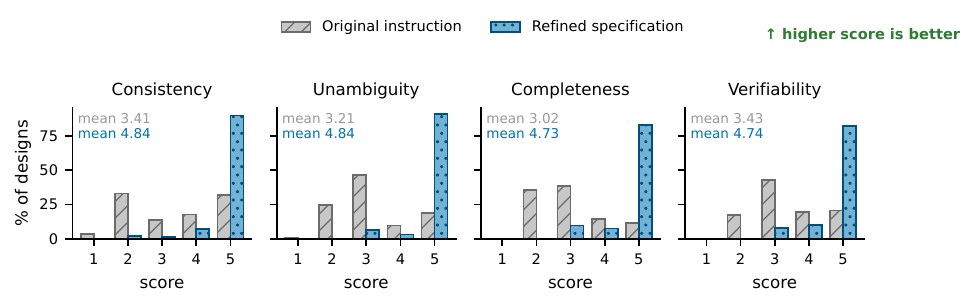}
  \caption{Distribution of the grades of the original instruction and the accepted refined specification for 4,446 designs that were graded again after rewriting.}
  \label{fig:quality}
\end{figure}

Of the 14,385 designs, 57 stop at Level~2 as misunderstood and 36 never obtain a refined specification that passes the program checks, so 14,292 reach the round trip.
From the original instruction, the builder rebuilds 26.1\% of the 14,385 designs, and 36.7\% when the exact list of ports is added to the instruction.
From the refined specification, it rebuilds 74.0\% at the first try and 78.2\% after revision, a gain of 52 percentage points, and every design type gains about 50 points (Figure~\ref{fig:bytype}).
Because the same builder rebuilds most designs once their specification is refined, the low rate from the original instructions comes from the instructions, not from the builder.
On 1,460 randomly sampled designs, a commercial builder (Claude Haiku~5.5) and an open-source one (Qwen3-Coder-30B) rebuild almost the same share from the original instruction (27\% and 28\%), and both rebuild far more from the refined specification (82\% and 71\%; Figure~\ref{fig:roundtrip}b).
The grades of Level~2 rise accordingly, from averages of 3.0--3.4 for the original instructions to 4.7--4.8 for the accepted refined specifications of the 4,446 designs that were graded again (Figure~\ref{fig:quality}).
Completeness improves most, as 83\% of the refined specifications receive the top grade, against 12\% of the original instructions.

\begin{keypoint}
\textbf{Key finding:} The \emph{builder} LLM rebuilds the circuit from the original instruction for only 26.1\% of the 14,385 designs. The multi-agent, tool-verified refinement of VeriLadder raises this to 78.2\%, and every accepted refined specification is confirmed by a formal equivalence check. The gain also holds for an open-source builder chosen for this study, Qwen3-Coder-30B (28\% to 71\%). Most instructions of the original dataset therefore do not determine their circuit, so the pairs teach an inaccurate mapping from natural language to RTL.
\end{keypoint}

\paragraph{Refinement trajectories as teacher data}
\label{sec:trajectories}
Existing Verilog datasets teach a model to write code from a specification, but none teaches it to write the specification.
Level~3 produces this data at no extra cost. For each design, VeriLadder stores its \emph{refinement trajectory}, which contains every refined specification the refiner wrote and, for each failed one, the counterexample that rejected it.
This gives 11,247 trajectories that end in a verified specification (Figure~\ref{fig:roundtrip}a) and 9,719 failed specifications with their counterexamples.

\subsubsection{Length of the refined specifications}
\begin{figure}[t]
  \centering
  \includegraphics[width=\linewidth]{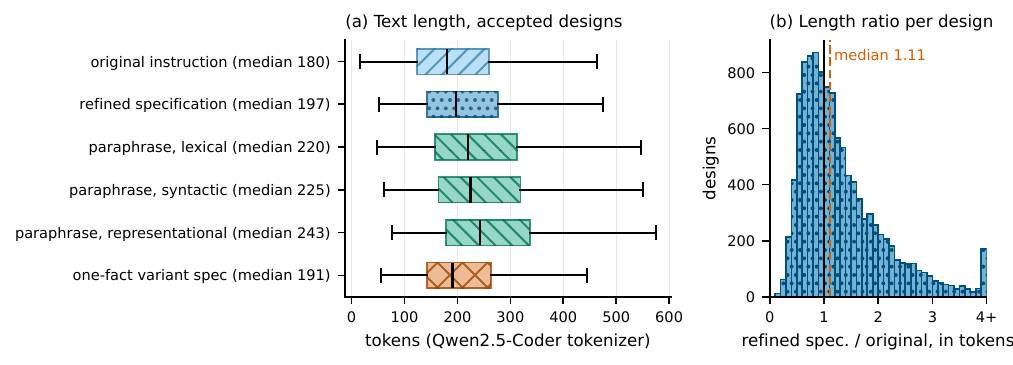}
  \caption{(a) Length in tokens of each kind of text for the 11,247 accepted designs (boxes: middle 50\%; lines: medians; whiskers: 1.5 times the box width). (b) Ratio of the length of each refined specification to the length of its original instruction.}
  \label{fig:tokens}
\end{figure}

A refined specification might be more precise only because it is longer, but it is not much longer (Figure~\ref{fig:tokens}a; Figure~\ref{fig:tokens}b shows the ratio of the two lengths for each design).
For the accepted designs, the median refined specification has 197 tokens and the median original instruction 180 (Qwen2.5-Coder tokenizer), and in words the two are almost equal, 128 and 129.
The few extra tokens come from exact names, widths and constants, not from more text.

%% ------------------------------------------------------------
\subsection{Level 4: Paraphrases}
\label{sec:level4}
Level~4 adds paraphrases of each accepted specification, each verified to describe the same circuit.
Figure~\ref{fig:ex-level4} shows three paraphrases of one specification.
% Real records of Levels 3-5 for RTLCoder id 25614 (adder, batch 16), a combinational saturating adder; it matches no
% VerilogEval or RTLLM problem (Rouge-L below the exclusion thresholds; neither benchmark has a saturating adder).
% Verbatim from spec_refinement/full_pilot/runs/portion16/results.jsonl: refined_rounds[0].spec (Behavior section) and
% its first counterexample trace, refined_rounds[1].spec (Behavior section), paraphrase.items[*].text (without the
% interface part), paraphrase.broken (text, the one-line change of mutated_code, mutant_vs_gold.trace, last step).
% Code from portions/p16/designs.jsonl (blank lines removed, module header joined on two lines). Only highlighting is added.
\begin{figure}[tp]
\centering
\scriptsize
\raggedright\noindent\textbf{Accepted specification of \texttt{adder}} (Figure~\ref{fig:ex-level3}b, a saturating adder; the interface part is omitted)\par\smallskip
\noindent
\begin{tcolorbox}[exbox, fontupper=\scriptsize, colback=green!4, colframe=green!45!black, width=\linewidth, title={Accepted specification}]
[...] Behavior: Compare the full unsigned mathematical sum A + B against 255. If the sum is greater than 255, Y is 255; otherwise, Y is the low 8 bits of the sum.
\end{tcolorbox}

\smallskip
\noindent\textbf{Three paraphrases}\par\smallskip
\noindent
\begin{tcolorbox}[exbox, fontupper=\scriptsize, colback=violet!3, colframe=violet!60!black, width=0.32\linewidth, title={Lexical}, equal height group=lv4]
\raggedright
Evaluate the complete unsigned mathematical total of \texttt{A} and \texttt{B} relative to 255. For a total exceeding 255, the output \texttt{Y} takes the value 255. For a total that does not exceed 255, \texttt{Y} carries the total's least significant 8 bits.
\tcblower
Rebuilt: \textbf{5 of 5} verified equivalent.
\end{tcolorbox}\hfill
\begin{tcolorbox}[exbox, fontupper=\scriptsize, colback=violet!3, colframe=violet!60!black, width=0.32\linewidth, title={Syntactic}, equal height group=lv4]
\raggedright
Determine the result by comparing the complete unsigned mathematical total of A and B with 0xFF: when that total exceeds 0xFF, the output Y takes the value 0xFF; when it does not exceed 0xFF, Y carries the total's least significant eight bits.
\tcblower
Rebuilt: \textbf{5 of 5} verified equivalent.
\end{tcolorbox}\hfill
\begin{tcolorbox}[exbox, fontupper=\scriptsize, colback=violet!3, colframe=violet!60!black, width=0.32\linewidth, title={Representational (table)}, equal height group=lv4]
\raggedright
Evaluate the complete unsigned mathematical total of \texttt{A} and \texttt{B} against 255. The output rule is:\par\smallskip
{\setlength{\tabcolsep}{3pt}\renewcommand{\arraystretch}{1.25}%
\begin{tabular}{@{}l>{\raggedright\arraybackslash}p{0.44\linewidth}@{}}
\toprule
\textbf{Total} & \textbf{\texttt{Y}} \\
\midrule
Greater than 255 & 255 \\
Not greater than 255 & The total's least significant 8 bits \\
\bottomrule
\end{tabular}}
\tcblower
Rebuilt: \textbf{5 of 5} verified equivalent.
\end{tcolorbox}
\caption{Real records of Level~4. Three paraphrases of the accepted specification of the saturating adder, one per kind of change (lexical, syntactic and representational). Each is rebuilt into a circuit verified equivalent to the code.}
\label{fig:ex-level4}
\end{figure}

\begin{figure}[t]
  \centering
  \includegraphics[width=\linewidth]{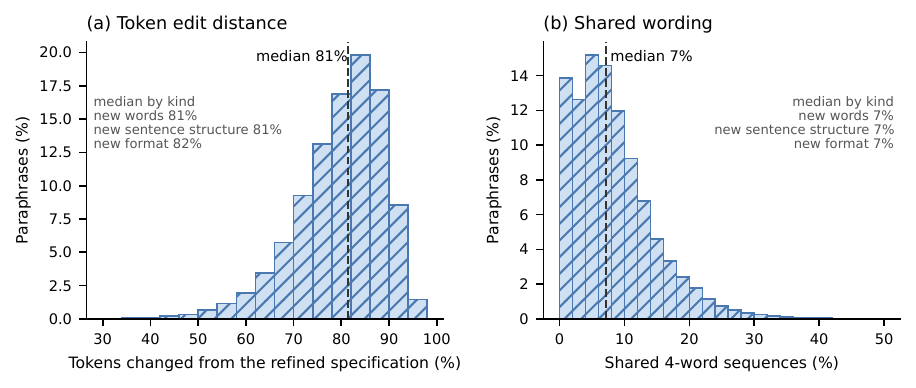}
  \caption{How much the 32,393 verified paraphrases differ from the refined specifications they reword. (a)~Share of tokens that must change to turn the refined specification into the paraphrase (token edit distance, Qwen2.5-Coder tokenizer). (b)~Share of the word 4-grams of a paraphrase that also appear in the refined specification. Dashed lines mark the medians; the medians of the three kinds of paraphrase are given in gray.}
  \label{fig:para-diff}
\end{figure}

\subsubsection{Motivation}
In real chip design work, different engineers ask for the same module in different words, and code-generation models are known to be sensitive to such changes~\cite{wang2023recode}.
A model that sees only one wording per design learns only one way of asking.
For Verilog, VeriAssist reworded the instructions of its own workflow, not the design descriptions, to test that its results do not depend on the wording~\cite{huang2024veriassist}, but no Verilog training set is known to contain paraphrases that are each verified to describe the same circuit.

\subsubsection{Procedure}
First, the refiner lists the \emph{critical facts} of the accepted refined specification. These are the statements an engineer needs to rebuild the circuit, one rule, constant, priority or timing property each, and each comes with the exact sentence of the specification that states it.
Then, shown as the \emph{Paraphraser} in Figure~\ref{fig:framework}, it writes three \emph{paraphrases}, specifications with the same meaning but different wording (Figure~\ref{fig:prompts}e).
Each one uses mainly one kind of change, which is
\emph{lexical} (other words for the same behavior, such as ``asserted low'' instead of ``active-low''),
\emph{syntactic} (a different sentence structure or order), or
\emph{representational} (another format, such as a table instead of prose, or hexadecimal instead of decimal numbers).
For every critical fact, the refiner must point to the exact words of the paraphrase that state it.
A program then checks each paraphrase. It must state every critical fact and name every port, and it must not contain code. It must also not copy long passages of the refined specification.
A paraphrase that passes these checks must then pass the round trip of Level~3 on its own.

\subsubsection{Results}
Paraphrases are somewhat longer than the refined specifications they reword (median 220--243 tokens; Figure~\ref{fig:tokens}a), because they often restate the facts as tables. Of 33,317 paraphrases that state all critical facts, 32,393 (97.2\%) pass the round trip. They are substantial rewordings. Turning a refined specification into one of its paraphrases changes a median of 81\% of its tokens (Figure~\ref{fig:para-diff}a), and of the four-word sequences (word 4-grams) in a paraphrase, a median of only 7\% also appear in the refined specification (Figure~\ref{fig:para-diff}b). The three kinds of paraphrase differ from the refined specification by about the same amount. Section~\ref{sec:train-level4} uses the paraphrases to test two LLMs.
Each accepted design therefore has up to four specifications, each verified to describe the same circuit.

\begin{keypoint}
\textbf{Key finding:} The paraphrases change a median of 81\% of the tokens of the refined specification and share only 7\% of its four-word sequences, yet 97.2\% of them still determine the same circuit. One verified design thus yields up to four specifications that differ substantially in wording but are verified to describe the same circuit. They can be used to train or test robustness to the wording of a request (Section~\ref{sec:train-level4}).
\end{keypoint}

%% ------------------------------------------------------------
\subsection{Level 5: One-fact variants}
\label{sec:level5}
Level~5 adds one-fact variants, pairs of a specification and code that differ from the accepted design in exactly one behavior.
Figure~\ref{fig:ex-level5} shows a variant of the saturating adder.
% Real records of Levels 3-5 for RTLCoder id 25614 (adder, batch 16), a combinational saturating adder; it matches no
% VerilogEval or RTLLM problem (Rouge-L below the exclusion thresholds; neither benchmark has a saturating adder).
% Verbatim from spec_refinement/full_pilot/runs/portion16/results.jsonl: refined_rounds[0].spec (Behavior section) and
% its first counterexample trace, refined_rounds[1].spec (Behavior section), paraphrase.items[*].text (without the
% interface part), paraphrase.broken (text, the one-line change of mutated_code, mutant_vs_gold.trace, last step).
% Code from portions/p16/designs.jsonl (blank lines removed, module header joined on two lines). Only highlighting is added.
\begin{figure}[tp]
\centering
\scriptsize
\noindent\textbf{One-fact variant of \texttt{adder}} (Figure~\ref{fig:ex-level3}b; the saturation value changes from 255 to 254)\par\smallskip
\noindent
\begin{tcolorbox}[exbox, fontupper=\scriptsize, colback=red!3, colframe=red!60!black, width=0.36\linewidth, title={Changed specification}, equal height group=lv5]
[...] Behavior: Compare the full unsigned mathematical sum A + B against 255. If the sum is greater than 255, Y is {\sethlcolor{hlbad}\hl{254}}; otherwise, Y is the low 8 bits of the sum.
\tcblower
Rebuilt: \textbf{5 of 5} verified equivalent to the changed code, \textbf{0 of 5} to the original code.
\end{tcolorbox}\hfill
\begin{tcolorbox}[exbox, fontupper=\scriptsize, colback=red!3, colframe=red!60!black, width=0.27\linewidth, title={Changed code (one line)}, equal height group=lv5]
\begin{Verbatim}[fontsize=\fontsize{6.4}{7.6}\selectfont]
if (A + B > 255) begin
  Y <= 254;   // was: 255
end else ...
\end{Verbatim}
\tcblower
Synthesizable, and a hard negative for the original specification.
\end{tcolorbox}\hfill
\begin{tcolorbox}[exbox, fontupper=\scriptsize, colback=black!3, colframe=black!35, width=0.34\linewidth, title={Counterexample}, equal height group=lv5]
\centering
\begin{tabular}{@{}c c c c@{}}
A & B & \multicolumn{2}{c}{Y} \\
 & & orig. & variant \\
143 & 128 & 255 & \textbf{\color{red!70!black}254} \\
\end{tabular}
\tcblower
\raggedright
The two circuits agree on every input whose sum fits in 8 bits and differ on every input that saturates.
\end{tcolorbox}
\caption{A real record of Level~5. The saturation value of the adder changes from 255 to 254 in both the specification and the code, and the counterexample shows where the variant and the original differ.}
\label{fig:ex-level5}
\end{figure}

\subsubsection{Motivation}
Many design bugs are a single wrong detail, such as a wrong reset value, the wrong clock edge, or a counter that stops one step too early.
A pair of a correct design and such a \emph{hard negative}, which differs from it in exactly one behavior, shows which detail makes the behavior different.

\subsubsection{Procedure}
Let $S$ be the accepted refined specification of a design and $D$ its code. The \emph{Trickster} agent in Figure~\ref{fig:framework} takes one critical fact of $S$, such as the active level of a reset, a priority, a constant, a delay or a clock edge, and changes it, for example to a saturation value of 254 instead of 255 (Figure~\ref{fig:prompts}f). It rewrites only the sentences of $S$ that state this fact, which gives the variant specification $S'$, and changes $D$ to match, which gives the variant code $D'$. The pair $(S', D')$ is a \emph{one-fact variant}.
A variant is kept only if two checks pass. First, the equivalence checker must find a counterexample, an input on which $D'$ and $D$ give different outputs, so they are really different circuits. Second, $S'$ must determine $D'$, which is tested with the round trip, in which the builder reads only $S'$ and at least one of its five attempts must be verified equivalent to $D'$. Figure~\ref{fig:ex-level5} shows both checks for the saturating adder.

\subsubsection{Results}
This gives 9,306 one-fact variants whose code differs from the original, of which 8,397 also pass the round trip. A variant specification has about the same length as the refined specification it changes (median 191 tokens; Figure~\ref{fig:tokens}a).
When the builder reads $S'$, it writes $D'$ in 86.5\% of attempts and $D$ in only 6.4\% (Section~\ref{sec:train-level5} compares this with an open-source model).
This also shows that the builder reads the specification instead of recalling the original code.
Together, the levels give each verified design a group of related examples. It contains $S$ and its paraphrases with the correct code $D$, and the variant code $D'$, which is wrong for all of these specifications in exactly one behavior, together with the specification $S'$ that it does satisfy (8,345 designs have such a group, with a verified variant and at least one paraphrase).
These one-fact variants are hard negatives for preference optimization and contrastive learning.

\begin{keypoint}
\textbf{Key finding:} A change of one requirement, such as a clock edge or a reset value, is followed by the builder in 87\% of attempts, and each one-fact variant is shown by a counterexample to describe a different circuit. The one-fact variants are therefore hard negatives that differ from a correct example in a single fact, and they reveal whether a model follows every fact of a specification (Section~\ref{sec:train-level5}).
\end{keypoint}

\section{Dataset Statistics}
\label{sec:stats}

This section describes the two source datasets and shows which designs each level keeps or removes.
The released dataset contains the labels of all levels for both datasets, the accepted and rejected refined specifications with their paraphrases and one-fact variants, and the refinement trajectories of Level~3. In total, the accepted designs carry 2.6 million tokens of refined specifications, 8.7 million tokens of paraphrases and 1.8 million tokens of variant specifications.

\subsection{Composition and duplicates}
Two designs are called \emph{near duplicates} if the sets of five-token sequences of their code have a Jaccard similarity of 0.8 or more.
RTLCoder has few duplicates. Only 0.3\% of its code is an exact copy of another design, and 5.5\% of designs have a near duplicate.
PyraNet has 2.7\% exact and 19.3\% near duplicates, and 24.7\% of its rows have an empty description or the description ``0''.

\subsection{Why designs fail synthesis}
\begin{figure}[t]
  \centering
  \includegraphics[width=\linewidth]{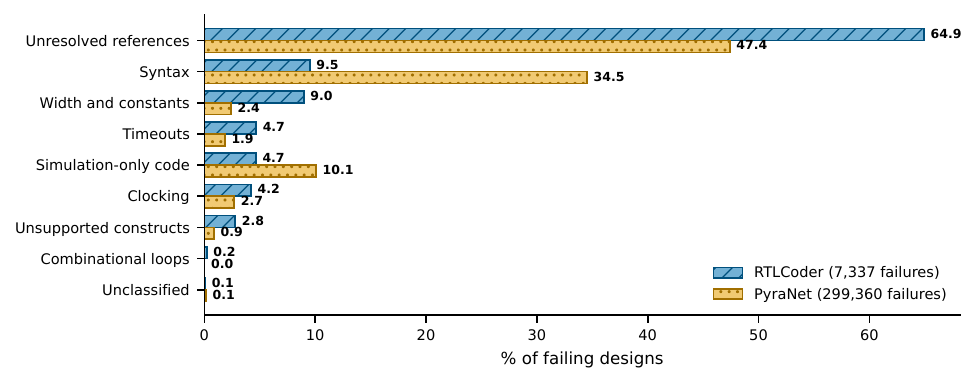}
  \caption{Failure families as a share of the designs that fail synthesis in each dataset.}
  \label{fig:families}
\end{figure}

The two datasets fail for different reasons (Figure~\ref{fig:families}).
In RTLCoder, 64.9\% of failures are unresolved references, mostly because the code uses a module that is not defined in the file.
In PyraNet, unresolved references are also the largest family (47.4\%), because many designs are parts of larger projects whose other files are missing; syntax errors follow with 34.5\%, and 10.1\% of failures are simulation-only commands such as \texttt{\$finish}, which come from testbench code stored as design code.

\subsection{Size of the designs}
\begin{figure}[t]
  \centering
  \includegraphics[width=\linewidth]{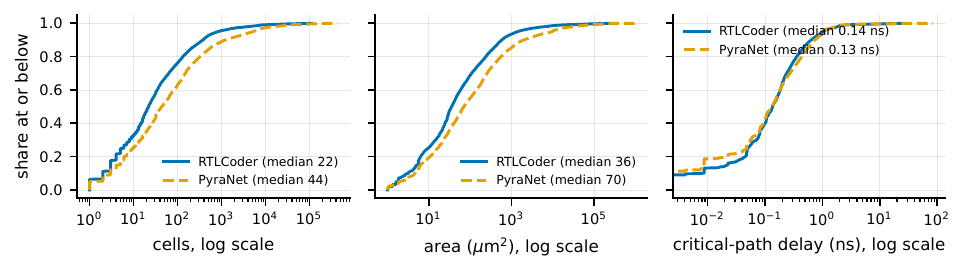}
  \caption{Cell count, area and critical-path delay of effectively synthesizable designs (Nangate 45\,nm). Each curve shows the share of designs at or below a value; the delay curve covers the designs with a plausible delay.}
  \label{fig:size}
\end{figure}

After mapping, PyraNet designs are about twice as large as RTLCoder designs (median 44 vs.\ 22 cells) and slightly deeper (median logic depth 4 vs.\ 3), and both datasets have the same share of sequential designs (57\% and 56\%; Figure~\ref{fig:size}).
The median critical-path delay is 0.14\,ns for RTLCoder and 0.13\,ns for PyraNet, which has more slow designs (90\% are faster than 0.64\,ns and 0.74\,ns). Delay and logic depth rank combinational designs almost the same way (Spearman 0.89 in both datasets) but sequential designs much less so (0.56 and 0.43), because the delay also includes the time a flip-flop needs to pass its value on and the extra delay of gates that drive many inputs, so both measures are released.

\subsection{Which designs are hard to specify}
\begin{figure}[t]
  \centering
  \includegraphics[width=\linewidth]{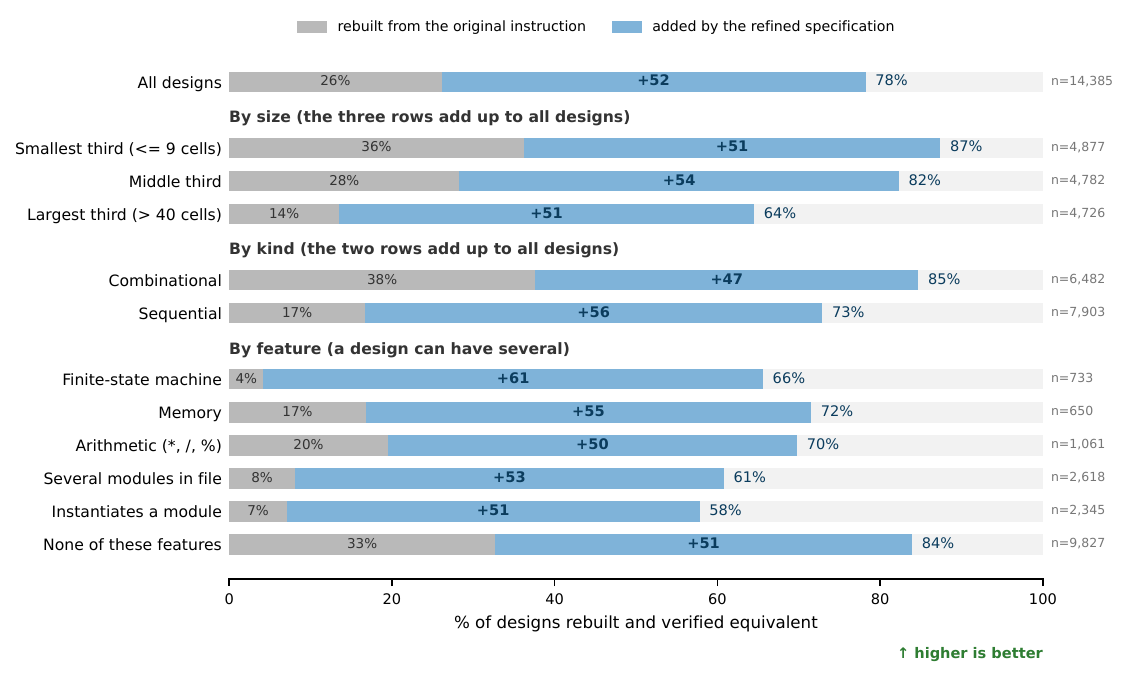}
  \caption{Share of designs rebuilt and verified equivalent, by design type. Gray is the share rebuilt from the original instruction, blue is what the refined specification adds (in percentage points), and the number at the end is the share rebuilt from the refined specification. The rows of each size and each kind add up to all 14,385 designs processed at Level~3; the feature rows overlap, since a design can have several features, and 9,827 designs have none of them.}
  \label{fig:bytype}
\end{figure}

The larger or more complex a design is, the less often its refined specification passes the round trip of Level~3 (Figure~\ref{fig:bytype}). Designs are grouped by size (cell count), by kind (with or without flip-flops), and by features of the code, namely a finite-state machine, a memory, arithmetic ($*$, $/$, $\%$), several modules in one file, or an instance of another module.
It falls from 87.4\% for the smallest third of designs to 64.5\% for the largest third, from 84.7\% for combinational to 72.9\% for sequential designs, and to 60.8\% for files that contain several modules.
Finite-state machines gain the most. Only 4.2\% are rebuilt from the original instruction, but 65.6\% from the refined specification.

\begin{keypoint}
\textbf{Key finding:} The original instructions most often fail to determine the circuit for large, sequential and multi-module designs and finite-state machines. Refinement raises the share of rebuilt designs in every group, and most for finite-state machines, from 4.2\% to 65.6\%, so it matters most for the designs that are hardest to specify.
\end{keypoint}
Designs that fail Level~3 are larger than those that pass (median 47 vs.\ 16 cells), so the accepted set contains more small designs.
The rejected designs are released with their last refined specification and counterexample.

\subsection{Refined specifications describe behavior, not one implementation}
\label{sec:impl}
\begin{figure}[t]
  \centering
  \includegraphics[width=\linewidth]{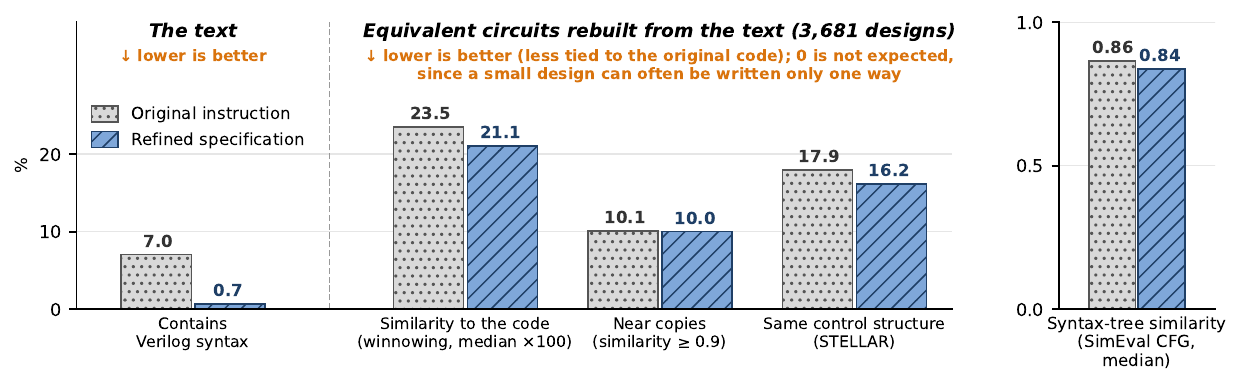}
  \caption{How closely the text and the circuits rebuilt from it follow the original code (lower is less tied to the code). ``The text'' covers all 11,247 accepted designs, and the rebuilds cover the designs rebuilt from both texts. The winnowing similarity is multiplied by 100, and SimEval (right) is on a scale from 0 to 1.}
  \label{fig:impl}
\end{figure}

A refined specification is accepted when the circuit built from it is verified equivalent to the original code.
A specification could pass this test by copying the code instead of describing what the circuit does, for example by restating the code line by line, naming its internal signals, or telling the builder which structure to use.
A model trained on such specifications would learn to reproduce one particular implementation, not to design a circuit from a description of its behavior.
Section~\ref{sec:level3} lists the guards against this during refinement, and the paragraphs below test, on all 11,247 accepted designs, whether they are enough (Figure~\ref{fig:impl}). If they are, the refined specification summarizes the code instead of translating it, it does not copy the code, and the circuits built from it differ in structure from the original code.

\begin{figure}[t]
  \centering
  \includegraphics[width=0.8\linewidth]{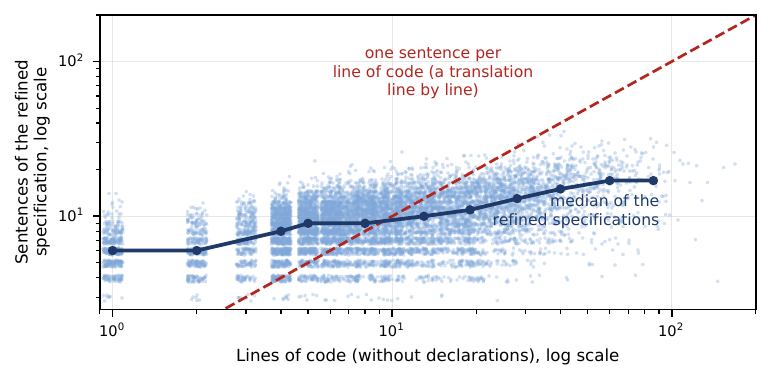}
  \caption{Length of each accepted refined specification against the size of its code (11,247 designs; points are jittered slightly). A translation of the code line by line would lie near the dashed line. The refined specifications grow much more slowly than the code, because they state the rule that the code implements instead of each of its statements.}
  \label{fig:spec-vs-code}
\end{figure}

\paragraph{The refined specification summarizes the code instead of translating it line by line.}
A translation of the code line by line would need about one sentence per line of code, and would name every internal signal.
Neither is the case (Figure~\ref{fig:spec-vs-code}).
The number of sentences grows only slowly with the size of the code and is roughly proportional to the 0.22nd power of the number of lines, where a translation line by line would give the first power.
For example, the 842 designs with 31 to 100 lines of code (not counting comments and declarations) have a median of 40 lines of code, while their refined specifications have a median of only 15 sentences. The 16 designs with more than 100 lines have a median of 110 lines of code and 20 sentences.
Small designs get more sentences than they have lines, because the refined specification also states the interface, the reset and the timing, which the code expresses in declarations.
On average, a refined specification names 28\% of the internal signals of its code, and in Figure~\ref{fig:impl-example} the 16 items of a case statement become two sentences that state the rule they implement.

\paragraph{The refined specification does not copy the code.}
Few refined specifications contain code. Only 0.7\% of them contain Verilog syntax, such as \texttt{always @} or \texttt{assign x =}, against 7.0\% of the original instructions.
Their wording is also their own. On average, only 0.25\% of their four-word sequences also appear in the code, against 0.83\% for the original instructions.
Refined specifications mention at least one internal signal in more designs than the original instructions do (41\% against 28\% of the designs that have internal signals), because some behavior, such as that of a clock divider, cannot be specified without a stored value. Most of these mentions are ordinary words such as \emph{state} or \emph{count}, and a name that looks like code appears for only 17\% of these designs.

\paragraph{The circuits built from the refined specification differ from the original code.}
The structural similarity of two pieces of code is measured with winnowing, the method of the plagiarism detector MOSS~\cite{schleimer2003winnowing}, in which names and numbers are replaced by placeholders, the port declarations are removed, and the similarity is the share of shared fingerprints of five-token sequences, from 0 (nothing shared) to 1 (the same token structure).
Over all 11,247 accepted designs, a rebuild from the refined specification that is verified equivalent to the original has a median similarity of only 0.24 to the original code, and only 6.0\% of the designs have a rebuild with a similarity of 0.9 or more.
A rebuild from the original instruction exists only for the 3,681 designs whose instruction is precise enough. On these designs, the rebuilds from the refined specification are slightly less similar to the original code than those from the instruction (median 0.21 against 0.23, Figure~\ref{fig:impl}), so refinement does not push the builder toward the original code.
The STELLAR fingerprint~\cite{rajabi2025stellar}, which compares the control structure of two pieces of code while ignoring names and values, gives the same result, with 16.2\% of the rebuilds from the refined specification and 17.9\% of those from the original instruction having exactly the control structure of the original code.
The syntax-tree similarity of SimEval~\cite{akyash2025simeval}, which compares the shapes of the syntax trees of two pieces of code, also agrees, with a median of 0.84 for the rebuilds from the refined specification against 0.86 for those from the original instruction.
Each measure is meant to compare the original instruction with the refined specification, and not to be compared with the other measures, since each one counts a different kind of match.
In 67\% of the designs, at least one rebuild makes different coding choices than the original, and rebuilds reuse on average only 23\% of the internal signal names of the original code.

\paragraph{The structure comes from the builder, not from the specification.}
The five rebuilds that one builder writes from the same refined specification are often nearly identical (median similarity 1.0), because a model has its own style.
A different builder writes different code from the same specification. For the 1,012 of the 1,460 randomly sampled designs that both builders rebuild from the final refined specification, the median similarity between a Qwen3-Coder-30B rebuild and a Claude Haiku~5.5 rebuild is 0.40, and both are verified equivalent to the original.
The same specification therefore admits several implementations, while the equivalence check ensures that they all have the same behavior.
Figure~\ref{fig:impl-example} shows an example. The original instruction prescribes a case statement, and every rebuild from it copies this structure; the refined specification states only the function, and the rebuilds from it that are verified equivalent compute it with a loop.

% Implementation independence example: RTLCoder id 25516 (batch 2); it matches no VerilogEval or RTLLM problem
% (checked by Rouge-L, port list and topic). Texts and code verbatim from spec_refinement/full_pilot/portions/p02/
% designs.jsonl and runs/portion02/results.jsonl (the instruction is shortened where marked [...]; the original code is
% condensed: header joined, 12 of the 16 case items replaced by ...). Only highlighting is added.
\begin{figure}[t]
\centering
\sethlcolor{hlmiss}
\footnotesize
\noindent
\begin{tcolorbox}[exbox, colback=black!3, colframe=black!35, width=0.49\linewidth, title={Original instruction (shortened)}, equal height group=im1]
[...] Write a Verilog module \texttt{MoleEncoder} that takes in a 16-bit binary number \texttt{mole16bit} and outputs a 5-bit binary number \texttt{mole\_position}. {\sethlcolor{hlbad}\hl{Your module should use a case statement to assign \texttt{mole\_position} based on the input \texttt{mole16bit}.}}
\tcblower
Rebuilt: \textbf{5 of 5}, all with a 16-entry case statement, like the original code.
\end{tcolorbox}\hfill
\begin{tcolorbox}[exbox, colback=black!3, colframe=black!35, width=0.49\linewidth, title={Original code (condensed)}, equal height group=im1]
\begin{Verbatim}[fontsize=\fontsize{6.4}{7.6}\selectfont]
module MoleEncoder(mole16bit, mole_position);
  input [15:0] mole16bit;
  output [4:0] mole_position;
  always @* begin
    case (mole16bit)
      16'b0000_0000_0000_0001: mole_position = 5'b00000;
      16'b0000_0000_0000_0010: mole_position = 5'b00001;
      ...
      16'b0100_0000_0000_0000: mole_position = 5'b01110;
      16'b1000_0000_0000_0000: mole_position = 5'b01111;
      default: mole_position = 5'b00000;
    endcase
  end
endmodule
\end{Verbatim}
\end{tcolorbox}

\medskip
\noindent
\begin{tcolorbox}[exbox, colback=blue!3, colframe=blue!50!black, width=0.49\linewidth, title={Refined specification (accepted)}, equal height group=im2]
Overview: MoleEncoder combinationally encodes a 16-bit one-hot input as the index of its set bit.
Interface: Input mole16bit is an unsigned 16-bit value. Output mole\_position is an unsigned 5-bit value.
Behavior: {\sethlcolor{hlfix}\hl{If mole16bit has exactly one set bit, mole\_position is the index of that bit}}, with bit 0 mapping to 0 and bit 15 mapping to 15. {\sethlcolor{hlfix}\hl{For every other input, including zero and inputs with multiple set bits, mole\_position is 0.}}
\end{tcolorbox}\hfill
\begin{tcolorbox}[exbox, colback=blue!3, colframe=blue!50!black, width=0.49\linewidth, title={Rebuilt from the refined specification}, equal height group=im2]
\begin{Verbatim}[fontsize=\fontsize{6.4}{7.6}\selectfont]
module MoleEncoder (
    input  wire [15:0] mole16bit,
    output reg  [4:0]  mole_position
);
    integer i;
    reg [4:0] idx;
    reg       valid;
    reg [4:0] cnt;
    always @(*) begin
        idx   = 5'd0;
        cnt   = 5'd0;
        for (i = 0; i < 16; i = i + 1) begin
            if (mole16bit[i]) begin
                idx = i[4:0];
                cnt = cnt + 5'd1;
            end
        end
        valid = (cnt == 5'd1);
        if (valid)
            mole_position = idx;
        else
            mole_position = 5'd0;
    end
endmodule
\end{Verbatim}
\tcblower
\textbf{2 of 5} rebuilds proven equivalent to the original code, both written as this loop; structural similarity to the original code 0.01.
\end{tcolorbox}

\smallskip
\caption{A refined specification states behavior, not structure (RTLCoder design \texttt{MoleEncoder}). The original instruction prescribes {\sethlcolor{hlbad}\hl{how}} to build the circuit, and every rebuild from it copies the case statement of the original code. The refined specification states only {\sethlcolor{hlfix}\hl{what}} the output is. The builder then writes a loop that counts the set bits, which differs from the original code in structure and is proven to compute the same function.}
\label{fig:impl-example}
\end{figure}

\begin{keypoint}
\textbf{Key finding:} The refined specifications describe behavior, not the implementation they were written from. They summarize the code instead of translating it line by line (about 15 sentences for 40 lines of code), almost never contain code (0.7\%, against 7.0\% of the original instructions), and the circuits built from them are verified equivalent to the original yet differ from it in structure (median similarity 0.24). A model trained on them therefore learns the behavior that a circuit must have, rather than to reproduce one particular implementation.
\end{keypoint}

\section{Training and Evaluation with VeriLadder}
\label{sec:training}
This section shows one use of each level. Level~1 trains a model to write synthesizable code, Level~2 to grade specifications, and Level~3 to write functionally correct code. Levels~4 and~5 test whether an LLM is robust to the wording of a request and whether it follows every fact of a specification.

\subsection{Level 1: synthesizability labels with KTO}
\label{sec:train-level1}
The RTLCoder designs are split, stratified by the Level~1 label, into 10,532 training prompts and 7,900 test prompts, and the rest are held out for validation. A synthesizable design is labeled desirable and a design that fails synthesis undesirable.
Eight open-source LLMs, from 1B to 32B parameters, are trained with Kahneman--Tversky optimization (KTO)~\cite{ethayarajh2024kto}, a preference method that needs only a desirable or undesirable label per example, not a synthesizable and a non-synthesizable answer to the same prompt. Each model is trained with low-rank adapters (LoRA~\cite{hu2022lora}; rank 16, $\alpha=32$, learning rate $5\times10^{-5}$, KTO $\beta=0.1$, effective batch size 8).
Each model then writes one design for every test prompt at temperature 0.1, and the design passes if Yosys synthesizes it without an error; as in the original labels, critical warnings and empty circuits are not counted as failures in this experiment.

Table~\ref{tab:train-level1} shows that training with the Level~1 labels raises the share of synthesizable code for every model, by 10.9 to 55.6 percentage points. With 7,900 prompts, the 95\% confidence interval of each share is about $\pm$0.9 points, so every gain is far larger than the uncertainty of the measurement.
The gain is largest for the models that start lowest, such as Llama-3.1-8B (from 16.9\% to 72.5\%), and after training a 1.5B model writes synthesizable code more often than the pre-trained 32B model.

\begin{table}[t]
\centering\small
\caption{Share of synthesizable code (\%) on the 7,900 test prompts, before and after KTO with the Level~1 labels (\hib{} higher is better).}
\label{tab:train-level1}
\begin{tabular}{lrrr}
\toprule
Model & Pre-trained\,\hib & KTO\,\hib & Gain\,\hib \\
\midrule
Qwen2.5-Coder-7B-Instruct & 77.5 & 89.0 & +11.5 \\
Qwen3-32B & 70.4 & 84.3 & +13.9 \\
Qwen2.5-7B-Instruct & 60.1 & 80.8 & +20.7 \\
Qwen2.5-Coder-1.5B-Instruct & 40.6 & 80.6 & +40.0 \\
Qwen2.5-14B-Instruct & 69.7 & 80.6 & +10.9 \\
Gemma-3-4B-it & 52.2 & 78.7 & +26.5 \\
Llama-3.1-8B-Instruct & 16.9 & 72.5 & +55.6 \\
Llama-3.2-1B-Instruct & 7.8 & 45.4 & +37.6 \\
\bottomrule
\end{tabular}
\end{table}

\subsection{Level 2: predicting the quality of a specification}
\label{sec:train-level2}
The Level~2 grades can train an open-source model to grade specifications without a commercial LLM. Of the 14,292 graded instructions, 300 are held out, and Qwen2.5-Coder-7B-Instruct is fine-tuned on the other 13,992 (LoRA rank 16, three epochs) to write the four grades as one JSON object.

\begin{table*}[t]
\centering\footnotesize\setlength{\tabcolsep}{4.5pt}
\caption{Functional pass@$k$ (\%) with five samples per problem (\hib{} higher is better). VeriLadder-Bench is the 300 held-out designs with refined specifications, checked by formal equivalence. Bold marks the best open-source model. The suffix -I denotes the Instruct version of a model.}
\label{tab:main-results}
\begin{tabular}{llcrrrrrr}
\toprule
 & & & \multicolumn{2}{c}{VerilogEval~v2~\cite{liu2023verilogeval}\,\hib} & \multicolumn{2}{c}{RTLLM~v2~\cite{lu2024rtllm}\,\hib} & \multicolumn{2}{c}{VeriLadder-Bench\,\hib} \\
\cmidrule(lr){4-5}\cmidrule(lr){6-7}\cmidrule(lr){8-9}
Type & Model & \shortstack{Open-\\source} & pass@1 & pass@5 & pass@1 & pass@5 & pass@1 & pass@5 \\
\midrule
\multirow{6}{*}{Foundation} & GPT-4o & \xmark & 65.9 & 71.2 & 57.6 & 66.0 & 86.5 & 92.7 \\
 & GPT-5 & \xmark & 71.2 & 80.8 & 57.6 & 66.0 & 87.6 & 94.7 \\
 & GPT-6-sol & \xmark & 85.5 & 89.1 & 71.2 & 76.0 & 92.9 & 93.7 \\
\cmidrule(l){2-9}
 & Llama-3.1-8B-I & \cmark & 23.0 & 34.9 & 16.7 & 32.0 & 41.5 & 69.7 \\
 & Qwen2.5-7B-I & \cmark & 25.5 & 38.7 & 26.2 & 43.6 & 55.4 & 75.4 \\
 & Qwen2.5-Coder-7B-I & \cmark & 31.7 & 47.3 & 35.0 & 50.0 & 61.7 & 79.5 \\
\midrule
\multirow{2}{*}{Specialized} & CodeV-CL-7B~\cite{zhao2026codev} & \cmark & 25.5 & 43.1 & 28.5 & 45.6 & 45.8 & 72.7 \\
 & RTLCoder-DS-6.7B~\cite{liu2024rtlcoder} & \cmark & 34.6 & 47.3 & 36.6 & 48.4 & 61.3 & 80.1 \\
\midrule
\multirow{2}{*}{\shortstack[l]{Trained on RTLCoder\\(original dataset)}} & Qwen2.5-7B-I & \cmark & 35.1 & 44.1 & 24.0 & 36.0 & 61.1 & 77.1 \\
 & Qwen2.5-Coder-7B-I & \cmark & 39.4 & 46.7 & 31.2 & 44.0 & 67.2 & 81.3 \\
\midrule
\rowcolor{gray!15} & Qwen2.5-7B-I & \cmark & 44.5 & 59.0 & 36.4 & 48.0 & 79.5 & 90.0 \\
\rowcolor{gray!15}\multirow{-2}{*}{\shortstack[l]{\textbf{Ours}, trained on\\VeriLadder Spec.}} & Qwen2.5-Coder-7B-I & \cmark & \textbf{50.0} & \textbf{61.5} & \textbf{45.2} & \textbf{54.0} & \textbf{83.1} & \textbf{92.3} \\
\bottomrule
\end{tabular}
\end{table*}

\begin{figure}[t]
\centering
\includegraphics[width=\linewidth]{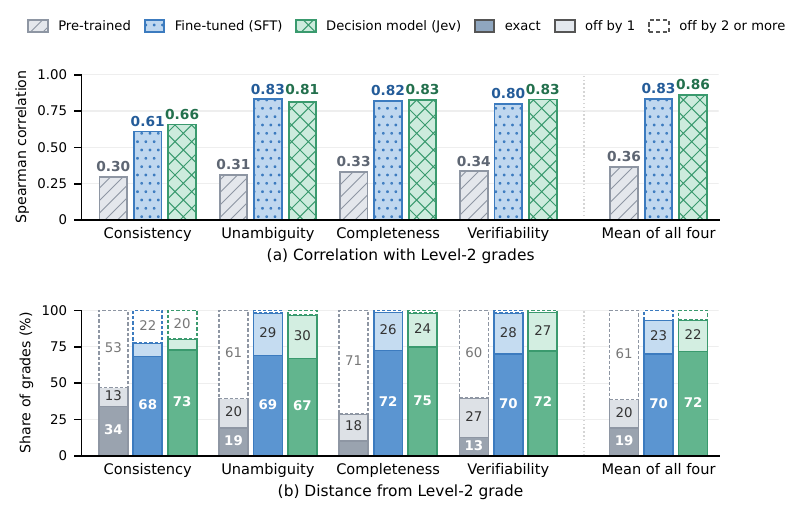}
\caption{Grades of Qwen2.5-Coder-7B-Instruct on the 300 held-out instructions, before fine-tuning, after fine-tuning to write the grades (SFT), and as a Jev-style decision model with a classification head (\hib{} higher is better). (a) Spearman correlation with the Level~2 grades. (b) Share of grades that are exact (dark), one point off (light) or two or more points off (dashed).}
\label{fig:train-level2}
\end{figure}

Figure~\ref{fig:train-level2} compares its grades with the held-out Level~2 grades, as correlations (a) and as the share of exact grades (b). Before fine-tuning, the model gives a 5 to almost every instruction, and its grades correlate only weakly with the Level~2 grades (Spearman 0.30--0.36). After fine-tuning, the correlation is 0.80--0.83 for unambiguity, completeness and verifiability, and 68--72\% of the grades are exact. Consistency is lower (0.61). The mean of the four grades also predicts the round-trip outcome with a ROC AUC of 0.86, close to the 0.90 that the Level~2 grades reach on the same 300 instructions and far above the 0.59 before fine-tuning.

A Jev-style decision model, trained on the same data, adds a classification head to the LLM instead of generating text. After one forward pass over the instruction, it answers each quality as a five-option multiple-choice question and returns a probability for every grade. It reaches about the same quality, with Spearman 0.66--0.83, 67--75\% exact grades and an AUC of 0.87, and its probabilities are better calibrated (expected calibration error 0.04 against 0.09).

\subsection{Level 3: functional correctness}
\label{sec:train-level3}
Level~3 pairs each accepted design with its original instruction and its refined specification, and both share the same verified code, so two models trained on them differ only in the text. Of the 11,247 accepted designs, 54 whose code has undriven nets, which an earlier version of the equivalence checker did not compare, are left out; 300 of the remaining 11,193 are held out as VeriLadder-Bench and 10,893 are used for training (LoRA rank 16, three epochs). Functional pass@$k$ is the probability that at least one of $k$ samples passes the testbench of the benchmark, estimated from five samples per problem~\cite{chen2021codex}.
Training on the refined specifications instead of the original instructions raises functional pass@1 from 39.4\% to 44.7\% on VerilogEval~v2 and from 31.2\% to 41.6\% on RTLLM~v2 for Qwen2.5-Coder-7B-Instruct, and from 35.1\% to 39.2\% and from 24.0\% to 33.6\% for Qwen2.5-7B-Instruct. Adding 3,248 designs of underrepresented categories from CodeV-R1~\cite{zhu2025codevr1} and OriGen~\cite{cui2024origen}, none of which matches a benchmark problem, gives the final models in Table~\ref{tab:main-results}. They clearly outperform their pre-trained versions, while the commercial models remain ahead. As in RTLCoder~\cite{liu2024rtlcoder} and DeepRTL~\cite{liu2025deeprtl}, the fine-tuned models are reported at their best temperature and the open-source baselines are averaged over the temperatures 0.2 to 1.0; averaged in the same way, the fine-tuned Coder model still reaches 47.8\% and 41.0\%.

% Main comparison. Sources: veriladder_train/evals/{api,api_val,sweep,sweep_q7,val*}/summary.json.
% Open-source baselines: mean over T = 0.2, 0.4, 0.6, 0.8, 1.0 (n = 5 samples per problem at each T).
% Ours (Level 3 + FSM and table augmentation, run "l3_haiku_fsmtab"): best T per column, chosen from T = 0.2, ..., 1.0 (n = 5)
% (Coder: 0.2 / 0.8 / 0.2 / 0.6; Qwen: 0.6 / 0.6 / 0.2 / 1.0). Mean over T = 0.2--1.0 for ours: Coder 47.8 / 59.6 / 41.0 / 50.4,
% Qwen 42.1 / 54.9 / 34.8 / 43.6.
% CodeV-CL-7B (yang-z/CodeV-CL-7B): Magicoder instruction template (its training format), mean over T 0.2-1.0, gen_rtlcoder.py --template magicoder.
% RTLCoder-DS-6.7B (ishorn5/RTLCoder-Deepseek-v1.1): same prompts as plain text, no chat template, mean over T 0.2-1.0 (local vLLM 0.18, gen_rtlcoder.py).
% Commercial: gpt-4o at T = 0.2, top_p 0.95; gpt-5 with reasoning effort minimal (it rejects none); gpt-6-sol with reasoning
% effort none; reasoning models at their fixed default temperature.
% VeriLadder-Bench: same temperature rule (Coder ours best T 0.4 / 0.6; mean over T 0.2-1.0 for ours 81.5 / 90.4). Qwen: ours best T 0.2 / 1.0 (mean over T 0.2-1.0: 78.3 / 88.0).

\subsection{Level 4: robustness to wording}
\label{sec:train-level4}
The paraphrases of Level~4 also serve as a test of how robust an LLM is to the wording of a prompt asking for RTL code. The test uses two builders, Claude Haiku~5.5, the builder of the pipeline, and Qwen3-Coder-30B, an open-source builder chosen for this study. On the same 1,460 randomly sampled designs as in Section~\ref{sec:level3}, each builder writes code from every verified paraphrase, and the equivalence checker compares each result with the original circuit (Figure~\ref{fig:robustness}).

For paraphrases, the kind of rewording barely matters. Haiku rebuilds the same circuit in 93.5--94.4\% of attempts for lexical, syntactic and representational paraphrases, and Qwen3-Coder in 69.9--72.2\% (Figure~\ref{fig:robustness}a), so neither model depends on one particular form of wording.

\subsection{Level 5: following every fact}
\label{sec:train-level5}
\begin{figure}[t]
  \centering
  \includegraphics[width=\linewidth]{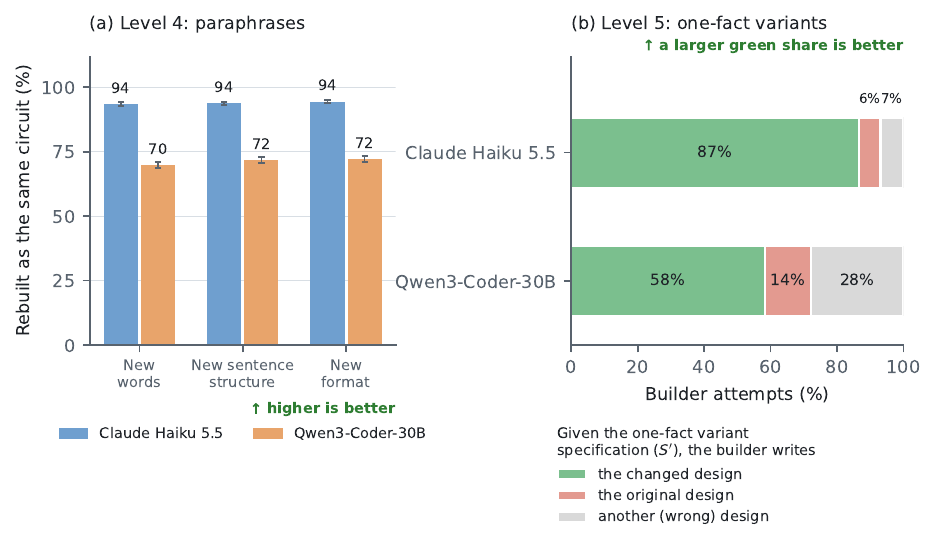}
  \caption{Levels 4 and 5 with two builders on 1,460 randomly sampled designs. (a)~Share of paraphrases rebuilt as the same circuit, by kind of paraphrase (95\% confidence intervals). (b)~Given the one-fact variant specification $S'$, the share of attempts in which the builder writes the changed design $D'$ (the intended outcome), the original design $D$, or another design that is wrong or does not compile.}
  \label{fig:robustness}
\end{figure}

The one-fact variants of Level~5 test whether an LLM follows every fact of a specification. With the same two builders and designs as in Section~\ref{sec:train-level4}, each builder reads the variant specification $S'$ of every variant (Section~\ref{sec:level5}) and writes code $C$. The equivalence checker then determines whether $C$ is equivalent to the variant code $D'$ (the model follows the change), to the original code $D$ (it falls back to the original design), or to neither. The two models differ clearly. Haiku follows the changed fact in 86.6\% of attempts and falls back to the original design in 6.4\%, while Qwen3-Coder follows it in only 58.2\% and falls back in 13.9\%, twice as often (Figure~\ref{fig:robustness}b). A one-fact variant thus exposes whether a model reads every fact of a specification or only recognizes the kind of circuit and writes its common version.

\section{Conclusion}
\label{sec:conclusion}
This paper presented VeriLadder, a Verilog dataset built as a ladder of five levels that follow the order in which a hardware engineer would examine a design.
Level~1 establishes that the code is synthesizable, Level~2 grades how well the original instruction describes the circuit, Level~3 rewrites the instruction into a refined specification that is verified to determine the circuit by a formal tool, Level~4 adds paraphrases that keep the same circuit, and Level~5 pairs each design with a near miss that differs in exactly one behavior.
Every check of correctness is decided by an EDA tool rather than by an LLM.

The released dataset applies the framework to 716,627 designs from two public corpora, labeling them for synthesizability, failure cause and post-synthesis metrics, grades the instructions of 14,292 designs, and provides 11,247 refined specifications verified to determine their circuit, together with 32,393 verified paraphrases, 8,397 one-fact variants and the refinement trajectories of 14,385 designs.
The analysis shows that most training instructions do not determine their circuit, that four standard qualities of a specification predict when they fail to do so, largely independent of the model that grades them, and that rewriting under a formal acceptance test resolves most of these cases.
The training experiments show that the levels can be used directly for training. The Level~1 labels raise the share of synthesizable code for all eight models tested, the Level~2 grades train an open-source grader whose grades correlate with them at a Spearman coefficient of up to 0.83, and fine-tuning Qwen2.5-Coder-7B-Instruct on the refined specifications raises its functional pass@1 from 31.7\% to 50.0\% on VerilogEval~v2 and from 35.0\% to 45.2\% on RTLLM~v2, well above the same model trained on the original instructions.
Beyond these uses, the paraphrases, one-fact variants and refinement trajectories provide training signals for robustness to wording, contrastive learning and specification refinement.

\bibliographystyle{ACM-Reference-Format}
\bibliography{refs}

% \appendix
% \input{sections/A_full_prompts}

\end{document}